\documentclass[11pt]{article}

\usepackage[final]{acl}

\usepackage{times}
\usepackage{latexsym}
\usepackage{paralist}

\usepackage[T1]{fontenc}

\usepackage[utf8]{inputenc}

\usepackage{microtype}

\usepackage{inconsolata}

\usepackage{graphicx}

\title{SPADER: Step-wise Peer Advantage with Diversity-Aware Exploration Rewards for Multi-Answer Question Answering}

\author{
 \textbf{Qiming Shi\textsuperscript{1,*}},
 \textbf{Zhaolu Kang\textsuperscript{2,*}},
 \textbf{Yunfan Zhou\textsuperscript{1}},
 \textbf{Di Weng\textsuperscript{3,$\dagger$}},
 \textbf{Yingcai Wu\textsuperscript{1}}
 \\ 
\textsuperscript{1}State Key Lab of CAD\&CG, Zhejiang University\\
\textsuperscript{2}School of Software and Microelectronics, Peking University\\
\textsuperscript{3}School of Software Technology, Zhejiang University\\
\textsuperscript{*}Equal contribution \qquad \textsuperscript{$\dagger$}Corresponding author\\
\texttt{\{qimingshi,dweng,ycwu\}@zju.edu.cn}
}

\usepackage{amsmath}
\usepackage{amssymb}
\usepackage{booktabs}
\usepackage{multirow} 
\usepackage{arydshln}
\usepackage{tabularx} 
\usepackage{pgfplots}
\pgfplotsset{compat=newest}
\usepackage{xcolor}
\usepackage[most]{tcolorbox}
\usepackage{enumitem}
\usepackage{amsfonts}
\usepackage{bbm}
\usepackage{stfloats}
\usepackage{float}

\begin{document}
\maketitle

\begin{abstract}
Large language models are increasingly deployed as tool-augmented agents to acquire information beyond parametric knowledge.
While recent work has improved long-horizon tool-use reasoning, most approaches focus on tasks with a single correct answer.
In contrast, many real-world queries require discovering a comprehensive set of valid answers, a setting known as Multi-Answer QA.
This setting raises two challenges: fine-grained credit assignment over long search trajectories and reward alignment for sustained exploration beyond easy high-frequency entities.
We propose \textbf{SPADER}, a reinforcement learning framework for long-horizon tool use in Multi-Answer QA.
SPADER includes Step-wise Peer Advantage (SPA), a critic-free step-level credit assignment mechanism that aligns parallel trajectories by decision step and estimates advantages from peer returns.
It also includes a diversity-aware exploration reward that promotes long-tail entity discovery by upweighting rare findings and downweighting redundant ones.
Experiments on QAMPARI, Mintaka, WebQSP, and QUEST show that SPADER generally improves recall and overall F1
over prompting-based agents, outcome-supervised RL methods, and recent step-level supervision approaches.
Our code and model weights are available at \url{https://github.com/KhanCold/spader}.
\end{abstract}

\section{Introduction}
\begin{figure}[!t]
  \centering
  \includegraphics[width=0.99\columnwidth]{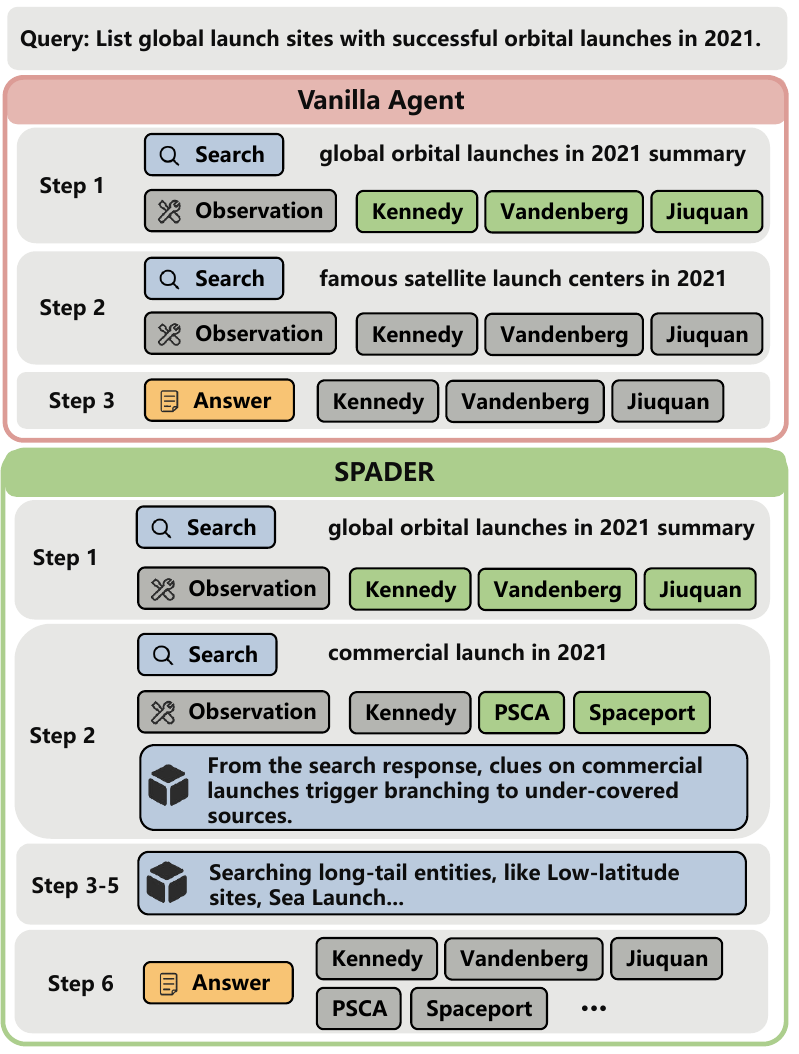}
  \caption{Vanilla Agent saturates on head entities and stops early, while SPADER keeps expanding toward long-tail launch sites under diversity-aware incentives.}
  \label{fig:intro_case}
\end{figure}

Large language models (LLMs) are increasingly deployed as agents that interact with external tools to acquire information beyond their parametric knowledge.
Recent work has extended retrieval-augmented generation~\cite{lewis2020retrieval} into long-horizon reasoning loops in which models iteratively issue search queries, inspect retrieved evidence, and decide when to terminate with a final answer~\cite{yao2022react,shao2023enhancing,trivedi2023interleaving}.
This paradigm transforms question answering from a single retrieval step into a sequential decision-making process, where an agent must decide at each step whether to issue further queries or terminate with a final answer.

Most existing work focuses on tasks that require only one correct answer.
In contrast, many real-world information needs are coverage-oriented, requiring systems to discover a comprehensive set of valid answers.
This setting, commonly referred to as Multi-Answer QA, shifts the central challenge from identifying a single correct answer to discovering a comprehensive set of valid answers.
For example, given the query \emph{``List global launch sites with successful orbital launches in 2021''} (Figure~\ref{fig:intro_case}), an agent should not stop at prominent entities like Kennedy Space Center.
Instead, it must sustain exploration to uncover valid, long-tail answers like PSCA and Kourou.

To achieve such sustained exploration, agents must learn to make multi-step decisions, including when to reformulate queries, which evidence to incorporate, and when to terminate~\cite{yao2022react,trivedi2023interleaving}.
Accordingly, recent work has increasingly framed tool-augmented QA as policy learning and optimized it with reinforcement learning (RL)~\cite{li2025search,jin2025search}.
However, in Multi-Answer QA, existing RL approaches for tool-augmented reasoning still face two key limitations.
First, long-horizon search trajectories make credit assignment difficult.
As agents execute a mixture of productive, redundant, and irrelevant searches over extended steps, identifying which step genuinely expands the answer coverage is challenging.
Actor-critic methods, such as PPO~\cite{schulman2017proximal}, depend on value networks whose estimation error tends to compound over long horizons, making credit assignment unreliable and training costly.
Critic-free methods, such as GRPO~\cite{shao2024deepseekmath}, improve stability, but their advantage estimation remains trajectory-level and thus provides only coarse credit assignment signal.
Approaches that provide step-level feedback~\cite{li2025r3rag,zheng2025stepsearch} often depend on external evaluators or process annotations, which limit scalability.

Second, commonly used reward formulations are poorly aligned with the objective of answer coverage.
For instance, F1-based rewards credit every matched entity equally, providing no additional incentive for retrieving difficult, long-tail answers.
Consequently, agents lack the motivation to sustain exploration for long-tail entities, resulting in premature halting after retrieving only prominent answers.

To address these challenges, we introduce \textbf{SPADER} (\textbf{S}tep-wise \textbf{P}eer \textbf{A}dvantage with \textbf{D}iversity-\textbf{A}ware \textbf{E}xploration \textbf{R}eward), an RL framework for long-horizon tool use in Multi-Answer QA.
SPADER combines two complementary ideas.
First, Step-wise Peer Advantage (SPA) provides a critic-free mechanism for fine-grained credit assignment.
Instead of evaluating trajectories only at the sequence level, SPA aligns parallel trajectories by decision step and estimates advantages using the empirical future-return distribution of peer trajectories at the same step.
This enables step-level policy optimization without requiring value networks or external evaluators.
Second, a diversity-aware exploration reward explicitly encourages long-tail entity discovery.
Beyond a base reward for retrieving valid entities, the reward for each entity is scaled inversely with its retrieval frequency across the trajectory group, encouraging exploration of new regions of the knowledge space while reducing redundant discoveries.

We evaluate SPADER on four Multi-Answer QA benchmarks: QAMPARI, Mintaka, WebQSP, and QUEST.
Experimental results show that SPADER consistently improves recall and overall F1 compared with prompting-based agents, outcome-supervised RL approaches, and recent step-level supervision methods.
These results highlight the importance of combining fine-grained credit assignment with diversity-aware exploration incentives when training long-horizon tool-use agents.

We summarize our contributions as follows:

\begin{compactitem}
    \item We propose Step-wise Peer Advantage (SPA), a critic-free step-level credit assignment mechanism that aligns parallel trajectories by decision step and derives advantages from peer trajectory return distributions.
    \item We introduce a diversity-aware exploration reward that encourages long-tail entity discovery by dynamically rewarding rare discoveries while reducing incentives for redundant high-frequency entities.
    \item Extensive experiments on four Multi-Answer QA benchmarks demonstrate that SPADER improves answer coverage and overall QA performance over strong prompting, outcome-supervised RL, and step-supervised baselines.
\end{compactitem}

\section{Related Work}
\subsection{Tool-Augmented QA Agents}

Early open-domain QA agents largely rely on single-step retrieval pipelines (e.g., RAG), which retrieve a fixed context before generation \cite{guu2020retrieval, lewis2020retrieval, karpukhin2020dense, xiong2020approximate, izacard2021leveraging, izacard2023atlas}.
These pipelines are effective for short factual queries but are less robust for iterative evidence gathering and multi-hop reasoning \cite{yang2018hotpotqa, trivedi2022musique}.
Recent work therefore studies tool-augmented LLM agents that interleave reasoning and retrieval across multiple steps, including ReAct, Iter-RetGen, IRCoT, and Search-o1 \cite{yao2022react, shao2023enhancing, trivedi2023interleaving, li2025search}.
These agentic pipelines generally improve performance on knowledge-intensive and multi-hop QA benchmarks \cite{yang2018hotpotqa, trivedi2022musique, izacard2023atlas}.

\subsection{Multi-Answer Question Answering}

While most QA research targets single-answer scenarios, Multi-Answer QA \cite{amouyal2023qampari, malaviya2023quest, sen2022mintaka, yih2016value} requires models to discover a comprehensive set of valid answers.
Prior works attempt to improve answer coverage through various pipeline enhancements, including query decomposition \cite{min2020ambigqa, perez2020unsupervised, khot2022decompose}, iterative query expansion \cite{khattab2021baleen}, and diverse passage reading \cite{qi2021answering, asai2022evidentiality}.
However, these methods largely rely on heuristic combinations of static retrievers and readers, lacking dynamic mechanisms to systematically explore the long-tail search space.

Recent RL-driven reasoning agents further extend tool-augmented QA by learning autonomous policies for retrieval and reasoning. Systems such as DeepRAG \cite{guan2025deeprag}, R1-Searcher \cite{song2025r1}, Search-r1 \cite{jin2025search}, and DeepResearcher \cite{zheng2025deepresearcher} demonstrate strong performance on complex QA.
However, standard reward formulations in these systems typically assign equal value to every correct entity found regardless of their retrieval difficulty.
As a result, existing agents may favor easy-to-retrieve entities and under-explore long-tail answers that remain discoverable.

\subsection{Fine-Grained Credit Assignment in RL}

Effective credit assignment in long-horizon reasoning requires attributing sparse rewards to intermediate steps \cite{sutton2018reinforcement, arjona2019rudder}.
Classical actor-critic training can be unstable or expensive in this regime due to value estimation over long trajectories \cite{schulman2017proximal, mnih2016asynchronous, haarnoja2018soft}.
Step-level supervision with PRMs and process-guided retrieval improves feedback granularity \cite{lightman2023let, uesato2022solving, li2025r3rag, zheng2025stepsearch}, but often requires costly annotation pipelines or strong external teachers.
In a complementary direction, critic-free methods such as GRPO estimate relative advantages from grouped trajectories \cite{shao2024deepseekmath, ahmadian2024back}, improving optimization efficiency but still aggregating credit at the trajectory level.
While recent works (e.g., GiGPO, SALT) achieve step-level credit assignment via exact state matching \cite{feng2025gigpo, li2025salt}, this strict requirement limits their applicability in Multi-Answer QA due to the inherently noisy and highly variable search observations. In contrast, our SPA avoids brittle state matching by directly aligning trajectories by decision step.

\section{Task Formulation}
We formulate the LLM-based long-horizon tool-use for Multi-Answer QA as a Markov Decision Process (MDP). Given a query $q$ and its ground-truth entity set $\mathcal{E}_{\mathrm{GT}}$, the agent interacts with the environment via multi-step search calls to output a complete and accurate entity list $\mathcal{E}_{\mathrm{final}}$.

At step $t$, the state $s_t {=} [q, a_1, o_1, \dots, a_{t-1}, o_{t-1}]$ concatenates $q$ and the interaction history, where $a_{t-1}$ is the previous action and $o_{t-1}$ is the textual observation returned by the search execution. Following policy $\pi_\theta(a_t|s_t)$, the agent selects an action from $\mathcal{A} = \{a_{\mathrm{search}}, a_{\mathrm{answer}}\}$:

\begin{compactitem}
    \item \textbf{Search Action} ($a_{\mathrm{search}}$): The agent generates a query to invoke an external search engine. The environment returns relevant document snippets as observation $o_t$, transitioning the state to $s_{t+1}$ to expand knowledge coverage without triggering the final answer generation.
    \item \textbf{Terminal Action} ($a_{\mathrm{answer}}$): Based on the accumulated state $s_t$, the agent outputs the final predicted entity set $\mathcal{E}_{\mathrm{final}}$. This acts as the absorbing state, immediately terminating the current trajectory.
\end{compactitem}

Thus, a full trajectory is strictly denoted as an alternating sequence $\tau = (s_1, a_1, o_1, \dots, s_L, a_L)$, where $L$ is the terminal step.

\section{SPADER}
\begin{figure*}[!t]
  \centering
  \includegraphics[width=0.94\textwidth]
  {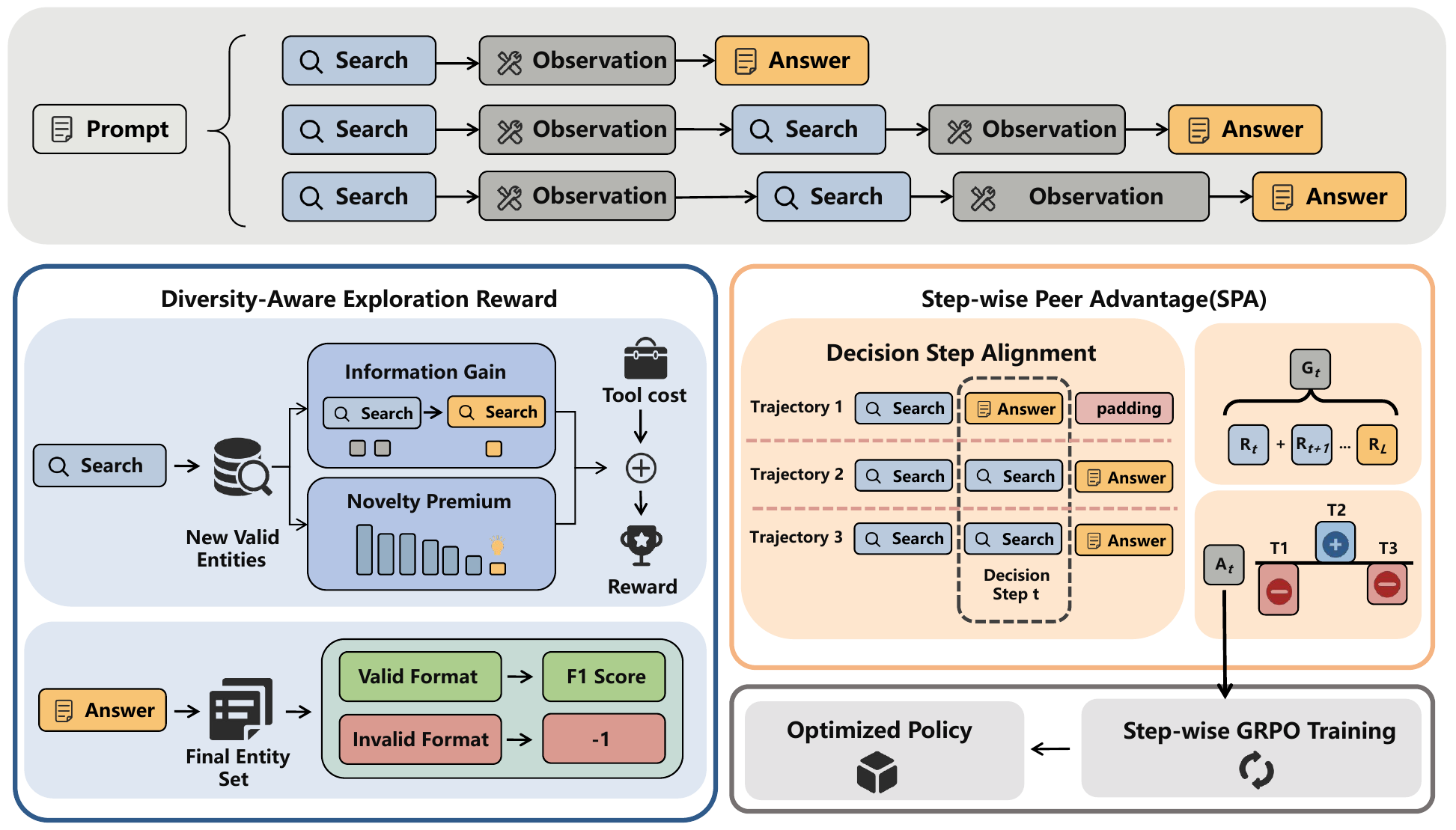}
    \caption{Overview of the SPADER framework. After sampling parallel trajectories, the framework evaluates each Search and Answer action using a Diversity-Aware Exploration Reward. Subsequently, Step-wise Peer Advantage (SPA) aligns these trajectories by decision steps to compute future returns ($G_t$) and step-wise advantages, which jointly optimize the policy through step-wise GRPO.}
\end{figure*}


To address the dual challenges of credit assignment and coverage-oriented exploration in long-horizon Multi-Answer QA, we propose SPADER (\textbf{S}tep-wise \textbf{P}eer \textbf{A}dvantage with \textbf{D}iversity-Aware \textbf{E}xploration \textbf{R}eward). Building on GRPO's group-based mechanism, SPADER shifts the advantage estimation from the standard trajectory level to a fine-grained, step-wise level.

Specifically, SPADER consists of two core components: a Step-wise Peer Advantage (SPA) for fine-grained credit assignment, and a Diversity-Aware Exploration Reward to prevent premature convergence to head entities.

\subsection{Step-wise Peer Advantage (SPA)}
\paragraph{Decision Step Alignment.}

To establish a baseline for step-level credit assignment without a parameterized value network, we align parallel trajectories by their \textit{decision step} $t$. 
Formally, a decision step $t$ corresponds to the agent emitting an action $a_t \in \{a_{\mathrm{search}}, a_{\mathrm{answer}}\}$ given the accumulated state $s_t$. 
Since trajectories sampled under the same prompt consume identical interaction turns at step $t$, aligning them establishes a prompt- and interaction-budget-conditioned baseline. 
This evaluates whether an action yields a higher future return than the group expectation under the exact same interaction budget, without assuming semantic or state equivalence across trajectories.

To handle varying lengths, we pad prematurely terminated trajectories to the maximum length of the group $L_{\max}^{(\mathrm{Group})}$.
They participate in the group baseline with a future return of $0$, while a Validity Mask $M_t^{(i)}$ blocks invalid gradient backpropagation after termination:
\begin{equation}
M_t^{(i)} = 
\begin{cases} 
1, & t \le L_i \text{ (Active)} \\ 
0, & t > L_i \text{ (Terminated)} 
\end{cases}
\end{equation}

\paragraph{Advantage Estimation.}

To achieve fine-grained step-level credit assignment, SPA leverages the future cumulative return distribution of the parallel trajectory group at the equivalent decision step as a dynamic baseline.

First, for the action taken by trajectory $i$ at step $t$, its actual future cumulative discounted return is calculated as $G_t^{(i)} = \sum_{k=0}^{L_i-t} \gamma^k r_{t+k}^{(i)}$, where $\gamma$ is the discount factor, and the summation continues until the actual termination step $L_i$.

Subsequently, we consider the $G$ trajectories sampled under the same prompt as a comparison group. 
At decision step $t$, the group baseline $\mu_t$ represents the expected empirical future return across the entire group: $\mu_t = \frac{1}{G}\sum_{j=1}^G \left( G_t^{(j)} \cdot M_t^{(j)} \right)$.

By comparing the individual return against the group empirical distribution and applying standardization, we obtain the step-wise relative advantage $\hat{A}_t^{(i)}$ for trajectory $i$ at step $t$:
\begin{equation}
\hat{A}_t^{(i)} = \frac{G_t^{(i)} - \mu_t}{\sigma_t} \cdot M_t^{(i)}
\end{equation}
where $\sigma_t$ is the standard deviation of the group returns at step $t$.

While standard Monte Carlo returns can exhibit high variance due to the compounding noise of future actions, SPA is designed to alleviate this issue through its step-aligned empirical baseline.
By subtracting the mean group return $\mu_t$ at the corresponding decision node, this formulation accounts for the shared expectation of future trajectory outcomes.

Such relative comparison yields a more localized estimate of the current action's utility, assigning positive advantages to steps that empirically outperform their peers.
Consequently, SPA facilitates fine-grained credit assignment for long-horizon search without necessitating parameterized value networks or external evaluators.

\paragraph{Training Objective.}

Our training objective adapts the GRPO framework~\cite{shao2024deepseekmath} to a step-wise setting.
By replacing the sequence-level advantage with our Step-wise Peer Advantage $\hat{A}_t^{(i)}$, we maximize the following objective:
\begin{multline}
\mathcal{J}(\theta) = \frac{1}{G} \sum_{i=1}^G \sum_{t=1}^{L_i} \bigg[ \min \Big( \rho_t^{(i)}(\theta) \hat{A}_t^{(i)}, \\
\operatorname{clip}\big(\rho_t^{(i)}(\theta), 1-\epsilon, 1+\epsilon\big) \hat{A}_t^{(i)} \Big) \\
- \beta_{KL} \mathbb{D}_{KL} \bigg]
\end{multline}
where $\rho_t^{(i)}(\theta) = \frac{P_\theta(y_t^{(i)}|x_t^{(i)})}{P_{\theta_{old}}(y_t^{(i)}|x_t^{(i)})}$ represents the probability ratio between the current and old policies.
$\mathbb{D}_{KL}$ denotes the KL divergence between the active and reference models to prevent excessive policy deviation, with $\beta_{KL}$ controlling the penalty strength.

\subsection{Diversity-Aware Exploration Reward}

To incentivize the discovery of long-tail entities and prevent premature convergence, we structure the multi-step reward into two distinct phases: a Dual-Axis exploration reward during the active search phase, and a global evaluation reward for the final generation phase.

\subsubsection{Dual-Axis Reward for Exploration}

For each search action ($a_t = a_{\mathrm{search}}$), we evaluate its exploration utility by decoupling the step-level reward into two orthogonal dimensions: horizontal information gain and vertical novelty.

For a newly discovered valid entity $e \in \mathcal{E}_{\mathrm{new}, t}^{(i)}$ acquired at step $t$ by trajectory $i$, its reward $R_{\mathrm{ent}}(e)$ is a weighted sum of two utilities:
\begin{equation}
R_{\mathrm{ent}}(e) = \alpha \cdot U_{\mathrm{gain}}(e) + \beta \cdot U_{\mathrm{novelty}}(e)
\end{equation}
where $\alpha, \beta > 0$ are balancing hyperparameters.

Specifically, the two utility components are defined as follows:
\begin{compactitem}
    \item \textbf{Horizontal Information Gain ($U_{\mathrm{gain}}$):} Represents the fundamental utility of retrieving a valid new entity, independent of peer behavior. We define this as a constant: $U_{\mathrm{gain}}(e) = 1$.
    \item \textbf{Vertical Novelty Premium ($U_{\mathrm{novelty}}$):} A reward dilution mechanism to penalize redundant discoveries across the parallel trajectory group $\mathcal{T}$. Let $N(e, \mathcal{T}) = \sum_{j=1}^G \mathbf{1}(e \in \tau_j)$ be the number of trajectories in the group retrieving $e$. The novelty utility is inversely proportional to its group-level frequency: $U_{\mathrm{novelty}}(e) = 1/N(e, \mathcal{T})$.
\end{compactitem}

This formulation establishes a dynamic incentive structure directly aligned with the coverage-oriented objective of Multi-Answer QA. Highly accessible ``head'' entities, which are easily discovered by the majority of parallel trajectories ($N \approx G$), face severe novelty dilution. Their marginal utility is consequently dominated merely by the base gain $\alpha$. Conversely, unique long-tail discoveries ($N=1$) yield the maximum reward $\alpha + \beta$. 

By explicitly tying the reward scale to an entity's group-level frequency, this mechanism naturally prevents the agent from greedily settling for high-frequency targets and terminating early. Instead, the policy retains a continuous incentive to offset the search costs and sustain exploration to uncover the comprehensive, long-tail answer distribution.

The overall step-level reward for trajectory $i$, upon executing the search action $a_{t}^{(i)}$, is aggregated as:
\begin{equation}
r_t^{(i)} = \frac{1}{|\mathcal{E}_{\mathrm{GT}}|} \bigg[ \sum_{e \in \mathcal{E}_{\mathrm{new}, t}^{(i)}} R_{\mathrm{ent}}(e) - \mathrm{Cost}(a_t^{(i)}) \bigg]
\end{equation}
where $\mathrm{Cost}(a_t^{(i)})$ imposes a fixed tool cost to discourage redundant or uninformative queries. Normalizing the total step reward by the ground-truth size $|\mathcal{E}_{\mathrm{GT}}|$ explicitly bounds the absolute reward scale, ensuring stable gradient updates during the RL process.

\subsubsection{Terminal Settlement Reward}
Upon executing the terminal action ($a_t^{(i)} = a_{\mathrm{answer}}$), the agent ceases search to output the final entity set $\mathcal{E}_{\mathrm{final}}^{(i)}$. We define the terminal reward $r_{L_i}^{(i)}$ as:

\begin{equation}
r_{L_i}^{(i)} = 
\begin{cases} 
-1, & \text{Invalid Format} \\ 
\operatorname{F1}(\mathcal{E}_{\mathrm{final}}^{(i)}, \mathcal{E}_{\mathrm{GT}}), & \text{Otherwise} 
\end{cases}
\end{equation}

The terminal reward $r_{L_i}^{(i)}$ enforces structural compliance and closes the task loop.
By providing a fixed evaluation of the final output, it serves as a necessary exit signal; when the expected novelty reward no longer outweighs the cumulative transition costs, this global constraint incentivizes the agent to cease exploration and deliver the final answer.

\section{Experiments}
\subsection{Datasets and Evaluation Metrics}
\newcolumntype{Y}{>{\centering\arraybackslash}X}
\begin{table*}[t]
\small
\centering
\renewcommand{\arraystretch}{1.4}
\tabcolsep=2pt 
\begin{tabularx}{\textwidth}{@{} l *{12}{Y} @{}}
\toprule
& \multicolumn{3}{c}{\textbf{QAMPARI}} & \multicolumn{3}{c}{\textbf{Mintaka}} & \multicolumn{3}{c}{\textbf{WebQSP}} & \multicolumn{3}{c}{\textbf{QUEST}} \\ \cmidrule(l){2-13} 
\multirow{-2}{*}{\textbf{Methods}} & P & R & F1 & P & R & F1 & P & R & F1 & P & R & F1 \\ \midrule
\multicolumn{13}{c}{\textbf{Llama3.1-8B}} \\ 
RAG        & 0.336 & 0.118 & 0.157 & 0.330 & 0.425 & 0.355 & 0.289 & 0.304 & 0.268 & 0.129 & 0.066 & 0.075 \\
ReAct      & 0.388 & 0.202 & 0.241 & 0.409 & 0.481 & 0.424 & 0.323 & 0.313 & 0.286 & 0.151 & 0.070 & 0.082 \\ \cdashline{1-13}
PPO        & 0.353 & 0.253 & 0.268 & 0.610 & 0.595 & 0.600 & 0.325 & 0.368 & 0.329 & 0.192 & 0.108 & 0.126 \\ 
GRPO       & 0.233 & 0.180 & 0.187 & 0.540 & 0.527 & 0.527 & 0.349 & 0.374 & 0.352 & 0.104 & 0.052 & 0.059 \\ \cdashline{1-13}
R3-RAG     & 0.386 & 0.290 & 0.303 & 0.617 & 0.630 & 0.607 & 0.452 & 0.413 & 0.403 & 0.169 & 0.117 & 0.121 \\ 
StepSearch & 0.418 & 0.258 & 0.304 & 0.639 & 0.612 & 0.618 & 0.499 & 0.387 & 0.403 & 0.181 & 0.104 & 0.124 \\ \cdashline{1-13}
SPADER (QAMPARI-only) & 0.420 & \textbf{0.411} & \textbf{0.394} & 0.538 & 0.594 & 0.558 & 0.427 & 0.392 & 0.378 & \textbf{0.216} & 0.112 & 0.128 \\
SPADER & \textbf{0.458} & 0.340 & 0.348 & \textbf{0.642} & \textbf{0.644} & \textbf{0.625} & \textbf{0.503} & \textbf{0.453} & \textbf{0.438} & 0.194 & \textbf{0.126} & \textbf{0.131} \\ \midrule
\multicolumn{13}{c}{\textbf{Qwen3-8B}} \\ 
RAG        & 0.181 & 0.119 & 0.128 & 0.297 & 0.448 & 0.325 & 0.314 & 0.284 & 0.273 & 0.154 & 0.065 & 0.078 \\
ReAct      & 0.419 & 0.173 & 0.218 & 0.384 & 0.466 & 0.404 & 0.344 & 0.351 & 0.318 & 0.158 & 0.077 & 0.090 \\ \cdashline{1-13}
PPO        & 0.410 & 0.256 & 0.279 & 0.573 & 0.546 & 0.550 & 0.380 & 0.340 & 0.336 & 0.176 & 0.100 & 0.109 \\ 
GRPO       & 0.423 & 0.182 & 0.226 & 0.528 & 0.497 & 0.504 & 0.339 & 0.290 & 0.290 & 0.167 & 0.092 & 0.104 \\ \cdashline{1-13}
R3-RAG     & 0.417 & 0.254 & 0.276 & 0.587 & 0.559 & 0.563 & 0.385 & 0.343 & 0.338 & 0.185 & 0.105 & 0.114 \\ 
StepSearch & 0.429 & 0.304 & 0.317 & 0.585 & 0.562 & 0.565 & 0.377 & 0.340 & 0.333 & 0.194 & \textbf{0.119} & 0.126 \\ \cdashline{1-13}
SPADER (QAMPARI-only) & \textbf{0.505} & \textbf{0.336} & \textbf{0.378} & 0.511 & 0.536 & 0.513 & 0.351 & 0.328 & 0.321 & 0.211 & 0.103 & 0.122 \\
SPADER & 0.474 & 0.312 & 0.343 & \textbf{0.625} & \textbf{0.626} & \textbf{0.611} & \textbf{0.423} & \textbf{0.409} & \textbf{0.388} & \textbf{0.233} & 0.109 & \textbf{0.127} \\ \bottomrule
\end{tabularx}
\caption{Performance comparison of our proposed approach against various baselines across four Multi-Answer QA datasets. We report Precision (P), Recall (R), and F1 scores for both Llama3.1-8B and Qwen3-8B backbones. Baseline methods are grouped into Prompting-based, Outcome-Supervised RL, and Step-level Supervision Evaluators.}
\label{tab:main-results}
\end{table*}

Our empirical benchmarks span four established Multi-Answer QA datasets: (1) QAMPARI \cite{amouyal2023qampari}, (2) Mintaka \cite{sen2022mintaka}, (3) WebQSP \cite{yih2016value}, and (4) QUEST \cite{malaviya2023quest}. 
More details about the datasets are provided in Appendix~\ref{sec:appendix_Datasets}.

For multi-answer evaluation, we compute instance-level Precision, Recall, and F1 based on the overlap between the set of unique predicted answers and the ground-truth answer set under exact match (EM) constraints. We report macro-averaged Precision, Recall, and F1 over all instances.

\subsection{Baselines}

To comprehensively evaluate our approach, we compare against several competitive baselines grouped into three established paradigms: 
(1) \textbf{Prompting-based Methods}: Approaches utilizing the frozen base model without policy optimization, including \textbf{RAG} \cite{lewis2020retrieval} and \textbf{ReAct} \cite{yao2022react}; 
(2) \textbf{Outcome-Supervised RL}: Frameworks optimizing sparse terminal rewards based on final correctness, comprising \textbf{PPO} \cite{schulman2017proximal} and \textbf{GRPO} \cite{shao2024deepseekmath}; and 
(3) \textbf{Step-level Supervision Evaluators}: Methods that explicitly score intermediate retrieval and reasoning steps, \textbf{R3-RAG}~\cite{li2025r3rag}, and \textbf{StepSearch}~\cite{zheng2025stepsearch}.

\subsection{Implementation Details}

We instantiate our proposed algorithm across two different models, utilizing Llama3.1-8B-Instruct \cite{llama} and Qwen3-8B \cite{qwen3} as the backbone language models.
For the search tool, we deploy a batched two-stage retrieval pipeline over the Wikipedia 2021 dump (aligning with the QAMPARI setup).
Upon receiving a batch of queries, the system processes each query individually by fetching the top-50 candidates via a local BM25 retriever \cite{rob2009bm25}, followed by re-ranking with BGE-Reranker-v2-M3 \cite{chen2024bge} to select the top-5 passages.

During the training phase, the models are trained on 12,000 instances sampled from the train splits: 6,000 from QAMPARI and 2,000 each from Mintaka, WebQSP, and QUEST.
To compute step-level rewards, after each rollout we normalize tool-response and ground-truth entities and then apply exact matching at each step; newly discovered entities at step $t$ are matched entities not seen in earlier steps.
The prompt template and further details are provided in Appendix~\ref{sec:appendix_implementation}.

\subsection{Main Results}

Table~\ref{tab:main-results} presents the evaluation results across the four Multi-Answer QA benchmarks. SPADER performs better than the compared baselines on both Llama3.1-8B and Qwen3-8B, achieving the best overall F1 scores.

\paragraph{Comparison with Prompting and Outcome-Supervised RL.} 
Prompting methods (RAG, ReAct) and outcome-supervised RL (PPO, GRPO) show lower performance, particularly in Recall. For instance, on the QAMPARI dataset (Llama3.1-8B), SPADER achieves an F1 of 0.348, yielding relative improvements of 29.9\% and 86.1\% over PPO and GRPO, respectively. These gaps suggest that step-wise credit assignment can be more effective for long-horizon search than relying only on terminal rewards or sequence-level returns.

\paragraph{Comparison with Step-Level Supervision.} 
When compared to methods utilizing step-level supervision (R3-RAG, StepSearch), SPADER also yields consistent improvements. This suggests that dynamically deriving advantages from intra-group empirical alignments (SPA) may provide a more calibrated learning signal than using external evaluators or static heuristics.

\paragraph{Model and Dataset Generalization.}
SPADER consistently achieves the highest F1 scores among all baselines on both Llama3.1-8B and Qwen3-8B across all datasets, showing robust, model-agnostic generalization across different architectures. Furthermore, \textbf{SPADER (QAMPARI-only)}, trained on 12,000 instances, remains competitive on unseen domains relative to prompting baselines.

\section{Analysis}
\subsection{Ablation study}

Table~\ref{tab:ablation-f1} evaluates the contribution of each SPADER component on Qwen3-8B.
Overall, removing any component degrades performance, indicating that the gains are not driven by a single design choice.

The w/o SPA setting causes a stable decline across datasets, supporting the role of Step-wise Peer Advantage in improving long-horizon action quality.
The w/o Novelty Premium setting leads to broad performance degradation, showing that the Vertical Novelty Premium is necessary for avoiding early collapse to frequent entities.

At the same time, the w/o Info Gain setting remains weak despite retaining the novelty term, with especially large drops on Mintaka and QAMPARI.
This pattern suggests that novelty alone can over-bias the policy toward long-tail entities when Horizontal Information Gain is removed.
Taken together, these results indicate that the Dual-Axis Reward for Exploration must be balanced, and the full SPADER objective is needed to achieve strong exploration quality without sacrificing final answer accuracy. Full Precision/Recall/F1 ablation results are provided in Appendix Table~\ref{tab:ablation-full}.

\begin{table}[H]
\small
\centering
\renewcommand{\arraystretch}{1.15}
\setlength{\tabcolsep}{5pt}
\begin{tabular}{lcccc}
\toprule
Methods & QAMPARI & Mintaka & WebQSP & QUEST \\
\midrule
SPADER & \textbf{0.343} & \textbf{0.611} & \textbf{0.388} & \textbf{0.127} \\
w/o SPA & 0.285 & 0.569 & 0.336 & 0.117 \\
w/o NP & 0.239 & 0.373 & 0.256 & 0.083 \\
w/o IG & 0.266 & 0.364 & 0.359 & 0.094 \\
\bottomrule
\end{tabular}
\caption{Ablation study on Qwen3-8B. We report F1 on four datasets. ``w/o'' denotes removing one component from SPADER. NP and IG denote novelty premium and information gain, respectively.}
\label{tab:ablation-f1}
\end{table}

\subsection{Hyperparameter Sensitivity Analysis}
We analyze the sensitivity of SPADER to the novelty weight $\beta$ in the dual-axis exploration reward.
All settings are fixed to the main configuration, and we sweep $\beta \in \{0, 1, 4, 8\}$ on Qwen3-8B with $\alpha=1$.
We do not additionally sweep $\alpha$.
In our reward design, $\alpha$ is a global scaling factor on the base gain term, while $\beta$ directly controls how strongly the policy prefers rare entities.

As shown in Table~\ref{tab:beta-sensitivity}, removing novelty reward ($\beta=0$) substantially hurts F1 on all datasets.
Increasing $\beta$ from $0$ to $4$ consistently improves performance, with the best results achieved at the main setting $\beta=4$.
When $\beta$ is further increased to $8$, 
performance plateaus and slightly declines,
suggesting that overly strong novelty pressure can bias exploration toward rare but less useful entities.

\begin{table}[H]
\small
\centering
\renewcommand{\arraystretch}{1.15}
\begin{tabular}{lcccc}
\toprule
$\beta$ & QAMPARI & Mintaka & WebQSP & QUEST \\
\midrule
0 & 0.239 & 0.373 & 0.256 & 0.083 \\
1 & 0.331 & 0.594 & 0.371 & 0.120 \\
4 & \textbf{0.343} & \textbf{0.611} & \textbf{0.388} & \textbf{0.127} \\
8 & 0.336 & 0.602 & 0.381 & 0.123 \\
\bottomrule
\end{tabular}
\caption{Sensitivity of SPADER to the novelty weight $\beta$ on Qwen3-8B. We report F1 on four datasets.}
\label{tab:beta-sensitivity}
\end{table}

\subsection{Search Efficiency Analysis}
To examine whether the performance gains come with excessive exploration, we analyze the average search counts required to retrieve a single valid entity.
The results on the Qwen3-8B are illustrated in Figure~\ref{fig:efficiency}.

While SPADER does not exhibit the lowest search counts per entity, it maintains competitive search efficiency while improving overall performance metrics (as shown in Table~\ref{tab:main-results}).
Methods with higher search counts exhibit different behaviors.
GRPO requires the highest search count per valid entity but yields poor recall, suggesting it struggles with redundant search calls. 

\begin{figure}[H]
    \centering
    \begin{tikzpicture}
        \begin{axis}[
            ybar,
            width=\columnwidth,
            height=5.5cm,
            bar width=16pt,
            enlarge x limits=0.15,
            ymin=0, ymax=5,
            ylabel={Avg. Search Count per Valid Entity},
            ylabel style={font=\small, yshift=-1ex},
            symbolic x coords={ReAct, R3-RAG, PPO, SPADER, StepSearch, GRPO},
            xtick=data,
            xticklabel style={rotate=30, anchor=north east, font=\small},
            nodes near coords,
            nodes near coords align={vertical},
            every node near coord/.append style={font=\footnotesize},
            axis lines*=left,
            ymajorgrids=true,
            grid style=dashed,
            ]
            
            \addplot[
                fill=blue!50!cyan, 
                draw=black,
                line width=0.5pt
            ] coordinates {
                (ReAct, 2.29)
                (R3-RAG, 2.47)
                (PPO, 2.58)
                (SPADER, 2.80)
                (StepSearch, 3.33)
                (GRPO, 4.44)
            };
        \end{axis}
    \end{tikzpicture}
    \caption{Search efficiency comparison on the Qwen3-8B backbone. The y-axis represents the average search calls needed to yield one valid entity.}
    \label{fig:efficiency}
\end{figure}

In contrast, baselines with lower search counts per entity, including ReAct, R3-RAG, and PPO, show lower recall.
This suggests that their seemingly high efficiency may partly come from earlier termination, where the agent captures head entities but explores less for long-tail ones.

\subsection{Temporal Exploration Dynamics}

To evaluate whether our approach effectively sustains exploration to uncover long-tail entities, we analyze the temporal discovery dynamics. Figure~\ref{fig:cumulative_discovery} tracks the cumulative number of distinct valid entities across interaction turns on the QAMPARI dataset using the Qwen3-8B backbone.

As Figure~\ref{fig:cumulative_discovery} shows, methods like GRPO and ReAct converge prematurely after the initial turn. While StepSearch extends its search into subsequent turns, SPADER maintains a higher cumulative yield across the entire trajectory. It captures more long-tail entities through a higher initial discovery and continued exploration in early turns, eventually plateauing as valid targets are exhausted.

\begin{figure}[H]
    \centering
    \begin{tikzpicture}
        \begin{axis}[
            width=0.98\columnwidth,
            height=6.0cm,
            xlabel={Interaction Turn ($t$)},
            ylabel={Cumulative Distinct Entities},
            xlabel style={font=\footnotesize},
            ylabel style={font=\footnotesize, yshift=-0.8ex},
            xmin=1, xmax=5,
            ymin=2, ymax=5.2, 
            xtick={1,2,3,4,5},
            legend columns=3,
            legend style={
                at={(0.5,1.02)},
                anchor=south,
                font=\scriptsize,
                draw=none,
                fill=none,
                /tikz/every even column/.append style={column sep=0.2cm}
            },
            axis lines*=left,
            axis line style={draw=black!65},
            tick style={black!55},
            tick label style={font=\footnotesize},
            ymajorgrids=true,
            grid style={line width=0.25pt, draw=gray!25},
            every axis plot/.append style={line width=1.3pt, mark options={scale=0.85, solid}}
        ]
        
        \addplot[color=gray!70!black, mark=square*] coordinates {
            (1, 2.19) (2, 2.24) (3, 2.24) (4, 2.24) (5, 2.24)
        }; \addlegendentry{ReAct}

        \addplot[color=orange!80!black, mark=triangle, densely dashed] coordinates {
            (1, 2.89) (2, 3.39) (3, 3.44) (4, 3.45) (5, 3.45)
        }; \addlegendentry{PPO}
        
        \addplot[color=orange!45!black, mark=triangle*] coordinates {
            (1, 2.35) (2, 2.38) (3, 2.38) (4, 2.38) (5, 2.38)
        }; \addlegendentry{GRPO}

        \addplot[color=blue!70!black, mark=diamond, densely dashed] coordinates {
            (1, 2.88) (2, 3.32) (3, 3.40) (4, 3.41) (5, 3.42)
        }; \addlegendentry{R3-RAG}
        
        \addplot[color=blue!45!black, mark=diamond*] coordinates {
            (1, 3.35) (2, 4.34) (3, 4.65) (4, 4.69) (5, 4.69)
        }; \addlegendentry{StepSearch}
        
        \addplot[color=teal!80!black, mark=*, line width=1.8pt] coordinates {
            (1, 3.75) (2, 4.76) (3, 4.87) (4, 4.87) (5, 4.87)
        }; \addlegendentry{\textbf{SPADER}}
        
        \end{axis}
    \end{tikzpicture}
    \caption{Cumulative distinct entities per interaction turn on QAMPARI.}
    \label{fig:cumulative_discovery}
\end{figure}
\subsection{Case Study}
We provide four qualitative cases in the Appendix~\ref{app:case_study} to complement the quantitative results, covering comparative evidence against baselines, behavior changes observed in ablations, and one failure case.

\section{Conclusion}

We study long-horizon tool use for Multi-Answer QA as a RL problem where the primary challenge lies in coverage-oriented exploration.
To address this, we propose SPADER, a framework combining Step-wise Peer Advantage (SPA) for critic-free step-level credit assignment with a diversity-aware exploration reward for long-tail entity discovery.
Experiments across four benchmarks show that SPADER effectively mitigates premature termination, consistently outperforming existing prompting-based and RL-driven agents.

\section*{Limitations}

Despite the effectiveness of SPADER, our study has three main limitations.
First, the experiments are conducted in an offline Wikipedia-based retrieval setting, which may not fully reflect the complexities of live web environments, such as content noise, API latency, and evolving knowledge.
Second, we evaluate our approach using 8B-parameter backbones on four QA benchmarks. Broader validation across larger model scales, multilingual contexts, and diverse tool-use tasks remains an important direction for future work.
Third, our reliance on coarse exact-match entity detection may miss lexical variants. Future extensions could incorporate task-specific canonicalization or LLM-based verification for more robust reward signals.

\section*{Ethical Consideration}
This work adheres to the ACL Ethics Policy. The benchmarks employed in our study are derived from publicly available and de-identified sources, ensuring no privacy risks or exposure of personally identifiable information. Regarding potential impact, although SPADER enhances answer coverage, the system may reflect intrinsic biases or factual inaccuracies present in the underlying language models and retrieved web corpora. We recommend exercising caution when deploying such agents in sensitive or high-stakes information-seeking domains.

\section*{Acknowledgement}
This work was supported by Ningbo Yongjiang Talent Programme (2024A-399-G). We sincerely appreciate the constructive feedback from the anonymous reviewers.

\bibliography{custom}

@inproceedings{amouyal2023qampari,
    title = {{QAMPARI}: A Benchmark for Open-domain Questions with Many Answers},
    author = {Amouyal, Samuel  and
      Wolfson, Tomer  and
      Rubin, Ohad  and
      Yoran, Ori  and
      Herzig, Jonathan  and
      Berant, Jonathan},
    booktitle = {Proceedings of the Third Workshop on Natural Language Generation, Evaluation, and Metrics (GEM)},
    month = dec,
    year = {2023},
    address = {Singapore},
    publisher = {Association for Computational Linguistics},
    url = {https://aclanthology.org/2023.gem-1.9/},
    pages = {97--110}
}

@inproceedings{sen2022mintaka,
    title = {Mintaka: A Complex, Natural, and Multilingual Dataset for End-to-End Question Answering},
    author = {Sen, Priyanka  and
      Aji, Alham Fikri  and
      Saffari, Amir},
    booktitle = {Proceedings of the 29th International Conference on Computational Linguistics},
    month = oct,
    year = {2022},
    address = {Gyeongju, Republic of Korea},
    publisher = {International Committee on Computational Linguistics},
    url = {https://aclanthology.org/2022.coling-1.138/},
    pages = {1604--1619}
}

@inproceedings{yih2016value,
    title = {The Value of Semantic Parse Labeling for Knowledge Base Question Answering},
    author = {Yih, Wen-tau  and
      Richardson, Matthew  and
      Meek, Chris  and
      Chang, Ming-Wei  and
      Suh, Jina},
    booktitle = {Proceedings of the 54th Annual Meeting of the Association for Computational Linguistics (Volume 2: Short Papers)},
    month = aug,
    year = {2016},
    address = {Berlin, Germany},
    publisher = {Association for Computational Linguistics},
    url = {https://aclanthology.org/P16-2033/},
    doi = {10.18653/v1/P16-2033},
    pages = {201--206}
}

@inproceedings{malaviya2023quest,
    title = {{QUEST}: A Retrieval Dataset of Entity-Seeking Queries with Implicit Set Operations},
    author = {Malaviya, Chaitanya  and
      Shaw, Peter  and
      Chang, Ming-Wei  and
      Lee, Kenton  and
      Toutanova, Kristina},
    booktitle = {Proceedings of the 61st Annual Meeting of the Association for Computational Linguistics (Volume 1: Long Papers)},
    month = jul,
    year = {2023},
    address = {Toronto, Canada},
    publisher = {Association for Computational Linguistics},
    url = {https://aclanthology.org/2023.acl-long.784/},
    doi = {10.18653/v1/2023.acl-long.784},
    pages = {14032--14047}
}

@inproceedings{yao2022react,
  title={{ReAct}: Synergizing Reasoning and Acting in Language Models},
  author={Shunyu Yao and Jeffrey Zhao and Dian Yu and Nan Du and Izhak Shafran and Karthik R Narasimhan and Yuan Cao},
  booktitle={The Eleventh International Conference on Learning Representations},
  year={2023},
  url={https://openreview.net/forum?id=WE_vluYUL-X}
}

@article{schulman2017proximal,
  title={Proximal policy optimization algorithms},
  author={Schulman, John and Wolski, Filip and Dhariwal, Prafulla and Radford, Alec and Klimov, Oleg},
  journal={arXiv preprint arXiv:1707.06347},
  year={2017}
}

@article{shao2024deepseekmath,
  title={Deepseekmath: Pushing the limits of mathematical reasoning in open language models},
  author={Shao, Zhihong and Wang, Peiyi and Zhu, Qihao and Xu, Runxin and Song, Junxiao and Bi, Xiao and Zhang, Haowei and Zhang, Mingchuan and Li, Y. K. and Wu, Yang and Guo, Daya},
  journal={arXiv preprint arXiv:2402.03300},
  year={2024},
  url={https://arxiv.org/abs/2402.03300}
}

@inproceedings{zheng2025stepsearch,
    title = {{S}tep{S}earch: Igniting {LLM}s Search Ability via Step-Wise Proximal Policy Optimization},
    author = {Zheng, Xuhui  and
      An, Kang  and
      Wang, Ziliang  and
      Wang, Yuhang  and
      Wu, Yichao},
    booktitle = {Proceedings of the 2025 Conference on Empirical Methods in Natural Language Processing},
    month = nov,
    year = {2025},
    address = {Suzhou, China},
    publisher = {Association for Computational Linguistics},
    url = {https://aclanthology.org/2025.emnlp-main.1106/},
    doi = {10.18653/v1/2025.emnlp-main.1106},
    pages = {21805--21830}
}

@inproceedings{li2025r3rag,
    title = {R3-{RAG}: Learning Step-by-Step Reasoning and Retrieval for {LLM}s via Reinforcement Learning},
    author = {Li, Yuan  and
      Luo, Qi  and
      Li, Xiaonan  and
      Li, Bufan  and
      Cheng, Qinyuan  and
      Wang, Bo  and
      Zheng, Yining  and
      Wang, Yuxin  and
      Yin, Zhangyue  and
      Qiu, Xipeng},
    booktitle = {Findings of the Association for Computational Linguistics: EMNLP 2025},
    month = nov,
    year = {2025},
    address = {Suzhou, China},
    publisher = {Association for Computational Linguistics},
    url = {https://aclanthology.org/2025.findings-emnlp.554/},
    doi = {10.18653/v1/2025.findings-emnlp.554},
    pages = {10491--10507},
}

@inproceedings{chen2024bge,
  title={{M}3-{E}mbedding: Multi-Linguality, Multi-Functionality, Multi-Granularity Text Embeddings Through Self-Knowledge Distillation},
  author={Chen, Jianlyu and Xiao, Shitao and Zhang, Peitian and Luo, Kun and Lian, Defu and Liu, Zheng},
  booktitle={Findings of the Association for Computational Linguistics: ACL 2024},
  month=aug,
  year={2024},
  address={Bangkok, Thailand},
  publisher={Association for Computational Linguistics},
  url={https://aclanthology.org/2024.findings-acl.137/},
  doi={10.18653/v1/2024.findings-acl.137},
  pages={2318--2335}
}

@article{rob2009bm25,
author = {Robertson, Stephen and Zaragoza, Hugo},
title = {The Probabilistic Relevance Framework: {BM25} and Beyond},
year = {2009},
issue_date = {April 2009},
publisher = {Now Publishers Inc.},
address = {Hanover, MA, USA},
volume = {3},
number = {4},
issn = {1554-0669},
url = {https://doi.org/10.1561/1500000019},
doi = {10.1561/1500000019},
journal = {Found. Trends Inf. Retr.},
month = apr,
pages = {333--389}
}

@InProceedings{guu2020retrieval,
  title = 	 {Retrieval Augmented Language Model Pre-Training},
  author =       {Guu, Kelvin and Lee, Kenton and Tung, Zora and Pasupat, Panupong and Chang, Mingwei},
  booktitle = 	 {Proceedings of the 37th International Conference on Machine Learning},
  pages = 	 {3929--3938},
  year = 	 {2020},
  volume = 	 {119},
  series = 	 {Proceedings of Machine Learning Research},
  month = 	 {13--18 Jul},
  publisher =    {PMLR},
  url = 	 {https://proceedings.mlr.press/v119/guu20a.html}
}

@inproceedings{lewis2020retrieval,
  title={Retrieval-Augmented Generation for Knowledge-Intensive {NLP} Tasks},
  author={Lewis, Patrick and Perez, Ethan and Piktus, Aleksandra and Petroni, Fabio and Karpukhin, Vladimir and Goyal, Naman and K\"{u}ttler, Heinrich and Lewis, Mike and Yih, Wen-tau and Rockt\"{a}schel, Tim and Riedel, Sebastian and Kiela, Douwe},
  booktitle={Advances in Neural Information Processing Systems},
  volume={33},
  publisher={Curran Associates, Inc.},
  url={https://proceedings.neurips.cc/paper_files/paper/2020/file/6b493230205f780e1bc26945df7481e5-Paper.pdf},
  pages={9459--9474},
  year={2020}
}

@inproceedings{shao2023enhancing,
    title = {Enhancing Retrieval-Augmented Large Language Models with Iterative Retrieval-Generation Synergy},
    author = {Shao, Zhihong  and
      Gong, Yeyun  and
      Shen, Yelong  and
      Huang, Minlie  and
      Duan, Nan  and
      Chen, Weizhu},
    booktitle = {Findings of the Association for Computational Linguistics: EMNLP 2023},
    month = dec,
    year = {2023},
    address = {Singapore},
    publisher = {Association for Computational Linguistics},
    url = {https://aclanthology.org/2023.findings-emnlp.620/},
    doi = {10.18653/v1/2023.findings-emnlp.620},
    pages = {9248--9274}
}

@inproceedings{trivedi2023interleaving,
    title = {Interleaving Retrieval with Chain-of-Thought Reasoning for Knowledge-Intensive Multi-Step Questions},
    author = {Trivedi, Harsh  and
      Balasubramanian, Niranjan  and
      Khot, Tushar  and
      Sabharwal, Ashish},
    booktitle = {Proceedings of the 61st Annual Meeting of the Association for Computational Linguistics (Volume 1: Long Papers)},
    month = jul,
    year = {2023},
    address = {Toronto, Canada},
    publisher = {Association for Computational Linguistics},
    url = {https://aclanthology.org/2023.acl-long.557/},
    doi = {10.18653/v1/2023.acl-long.557},
    pages = {10014--10037}
}

@inproceedings{guan2025deeprag,
  title={Deep{RAG}: Thinking to Retrieve Step by Step for Large Language Models},
  author={Guan, Xinyan and Zeng, Jiali and Meng, Fandong and Xin, Chunlei and Lu, Yaojie and Lin, Hongyu and Han, Xianpei and Sun, Le and Zhou, Jie},
  booktitle={The Fourteenth International Conference on Learning Representations},
  year={2026},
  url={https://openreview.net/forum?id=VI2YaggHIF}
}

@inproceedings{li2025search,
    title = {Search-o1: Agentic Search-Enhanced Large Reasoning Models},
    author = {Li, Xiaoxi  and
      Dong, Guanting  and
      Jin, Jiajie  and
      Zhang, Yuyao  and
      Zhou, Yujia  and
      Zhu, Yutao  and
      Zhang, Peitian  and
      Dou, Zhicheng},
    booktitle = {Proceedings of the 2025 Conference on Empirical Methods in Natural Language Processing},
    month = nov,
    year = {2025},
    address = {Suzhou, China},
    publisher = {Association for Computational Linguistics},
    url = {https://aclanthology.org/2025.emnlp-main.276/},
    doi = {10.18653/v1/2025.emnlp-main.276},
    pages = {5420--5438},
}

@article{song2025r1,
  title={R1-searcher: Incentivizing the search capability in llms via reinforcement learning},
  author={Song, Huatong and Jiang, Jinhao and Min, Yingqian and Chen, Jie and Chen, Zhipeng and Zhao, Wayne Xin and Fang, Lei and Wen, Ji-Rong},
  journal={arXiv preprint arXiv:2503.05592},
  year={2025}
}

@inproceedings{zheng2025deepresearcher,
    title = {{D}eep{R}esearcher: Scaling Deep Research via Reinforcement Learning in Real-world Environments},
    author = {Zheng, Yuxiang  and
      Fu, Dayuan  and
      Hu, Xiangkun  and
      Cai, Xiaojie  and
      Ye, Lyumanshan  and
      Lu, Pengrui  and
      Liu, Pengfei},
    booktitle = {Proceedings of the 2025 Conference on Empirical Methods in Natural Language Processing},
    month = nov,
    year = {2025},
    address = {Suzhou, China},
    publisher = {Association for Computational Linguistics},
    url = {https://aclanthology.org/2025.emnlp-main.22/},
    doi = {10.18653/v1/2025.emnlp-main.22},
    pages = {414--431},
}

@inproceedings{lightman2023let,
  title={Let's Verify Step by Step},
  author={Lightman, Hunter and Kosaraju, Vineet and Burda, Yuri and Edwards, Harrison and Baker, Bowen and Lee, Teddy and Leike, Jan and Schulman, John  and Sutskever, Ilya and Cobbe, Karl},
  booktitle={International Conference on Learning Representations},
  pages={39578--39601},
  url={https://proceedings.iclr.cc/paper_files/paper/2024/file/aca97732e30bcf1303bc22ac3924fd16-Paper-Conference.pdf},
  volume={2024},
  year={2024}
}

@inproceedings{izacard2021leveraging,
    title = {Leveraging Passage Retrieval with Generative Models for Open Domain Question Answering},
    author = {Izacard, Gautier  and
      Grave, Edouard},
    booktitle = {Proceedings of the 16th Conference of the European Chapter of the Association for Computational Linguistics: Main Volume},
    month = apr,
    year = {2021},
    address = {Online},
    publisher = {Association for Computational Linguistics},
    url = {https://aclanthology.org/2021.eacl-main.74/},
    doi = {10.18653/v1/2021.eacl-main.74},
    pages = {874--880}
}

@article{izacard2023atlas,
  title={Atlas: Few-shot learning with retrieval augmented language models},
  author={Izacard, Gautier and Lewis, Patrick and Lomeli, Maria and Hosseini, Lucas and Petroni, Fabio and Schick, Timo and Dwivedi-Yu, Jane and Joulin, Armand and Riedel, Sebastian and Grave, Edouard},
  journal={Journal of Machine Learning Research},
  volume={24},
  number={251},
  pages={1--43},
  year={2023}
}

@inproceedings{karpukhin2020dense,
    title = {Dense Passage Retrieval for Open-Domain Question Answering},
    author = {Karpukhin, Vladimir  and
      Oguz, Barlas  and
      Min, Sewon  and
      Lewis, Patrick  and
      Wu, Ledell  and
      Edunov, Sergey  and
      Chen, Danqi  and
      Yih, Wen-tau},
    booktitle = {Proceedings of the 2020 Conference on Empirical Methods in Natural Language Processing (EMNLP)},
    month = nov,
    year = {2020},
    address = {Online},
    publisher = {Association for Computational Linguistics},
    url = {https://aclanthology.org/2020.emnlp-main.550/},
    doi = {10.18653/v1/2020.emnlp-main.550},
    pages = {6769--6781}
}

@inproceedings{xiong2020approximate,
  title={Approximate Nearest Neighbor Negative Contrastive Learning for Dense Text Retrieval},
  author={Lee Xiong and Chenyan Xiong and Ye Li and Kwok-Fung Tang and Jialin Liu and Paul N. Bennett and Junaid Ahmed and Arnold Overwijk},
  booktitle={International Conference on Learning Representations},
  year={2021},
  url={https://openreview.net/forum?id=zeFrfgyZln}
}

@inproceedings{yang2018hotpotqa,
    title = {{H}otpot{QA}: A Dataset for Diverse, Explainable Multi-hop Question Answering},
    author = {Yang, Zhilin  and
      Qi, Peng  and
      Zhang, Saizheng  and
      Bengio, Yoshua  and
      Cohen, William  and
      Salakhutdinov, Ruslan  and
      Manning, Christopher D.},
    booktitle = {Proceedings of the 2018 Conference on Empirical Methods in Natural Language Processing},
    month = oct # "-" # nov,
    year = {2018},
    address = {Brussels, Belgium},
    publisher = {Association for Computational Linguistics},
    url = {https://aclanthology.org/D18-1259/},
    doi = {10.18653/v1/D18-1259},
    pages = {2369--2380}
}

@article{trivedi2022musique,
    title = {{M}u{S}i{Q}ue: Multihop Questions via Single-hop Question Composition},
    author = {Trivedi, Harsh  and
      Balasubramanian, Niranjan  and
      Khot, Tushar  and
      Sabharwal, Ashish},
    journal = {Transactions of the Association for Computational Linguistics},
    volume = {10},
    year = {2022},
    address = {Cambridge, MA},
    publisher = {MIT Press},
    url = {https://aclanthology.org/2022.tacl-1.31/},
    doi = {10.1162/tacl_a_00475},
    pages = {539--554},
}

@inproceedings{min2020ambigqa,
    title = {{A}mbig{QA}: Answering Ambiguous Open-domain Questions},
    author = {Min, Sewon  and
      Michael, Julian  and
      Hajishirzi, Hannaneh  and
      Zettlemoyer, Luke},
    booktitle = {Proceedings of the 2020 Conference on Empirical Methods in Natural Language Processing (EMNLP)},
    month = nov,
    year = {2020},
    address = {Online},
    publisher = {Association for Computational Linguistics},
    url = {https://aclanthology.org/2020.emnlp-main.466/},
    doi = {10.18653/v1/2020.emnlp-main.466},
    pages = {5783--5797}
}

@article{uesato2022solving,
  title={Solving math word problems with process-and outcome-based feedback},
  author={Uesato, Jonathan and Kushman, Nate and Kumar, Ramana and Song, Francis and Siegel, Noah and Wang, Lisa and Creswell, Antonia and Irving, Geoffrey and Higgins, Irina},
  journal={arXiv preprint arXiv:2211.14275},
  year={2022}
}

@inproceedings{sheng2024hybridflow,
author = {Sheng, Guangming and Zhang, Chi and Ye, Zilingfeng and Wu, Xibin and Zhang, Wang and Zhang, Ru and Peng, Yanghua and Lin, Haibin and Wu, Chuan},
title = {{HybridFlow}: A Flexible and Efficient {RLHF} Framework},
year = {2025},
publisher = {Association for Computing Machinery},
address = {New York, NY, USA},
url = {https://doi.org/10.1145/3689031.3696075},
doi = {10.1145/3689031.3696075},
booktitle = {Proceedings of the Twentieth European Conference on Computer Systems},
pages = {1279--1297},
series = {EuroSys '25}
}

@article{qwen3,
    title={{Qwen3} Technical Report},
    author={An Yang and Anfeng Li and Baosong Yang and Beichen Zhang and Binyuan Hui and Bo Zheng and Bowen Yu and Chang Gao and Chengen Huang and Chenxu Lv and Chujie Zheng and Dayiheng Liu and Fan Zhou and Fei Huang and Feng Hu and Hao Ge and Haoran Wei and Huan Lin and Jialong Tang and Jian Yang and Jianhong Tu and Jianwei Zhang and Jianxin Yang and Jiaxi Yang and Jing Zhou and Jingren Zhou and Junyang Lin and Kai Dang and Keqin Bao and Kexin Yang and Le Yu and Lianghao Deng and Mei Li and Mingfeng Xue and Mingze Li and Pei Zhang and Peng Wang and Qin Zhu and Rui Men and Ruize Gao and Shixuan Liu and Shuang Luo and Tianhao Li and Tianyi Tang and Wenbiao Yin and Xingzhang Ren and Xinyu Wang and Xinyu Zhang and Xuancheng Ren and Yang Fan and Yang Su and Yichang Zhang and Yinger Zhang and Yu Wan and Yuqiong Liu and Zekun Wang and Zeyu Cui and Zhenru Zhang and Zhipeng Zhou and Zihan Qiu},
    journal = {arXiv preprint arXiv:2505.09388},
    year={2025}
}

@article{llama,
  title={The {Llama} 3 Herd of Models},
  author={Aaron Grattafiori and Abhimanyu Dubey and Abhinav Jauhri and Abhinav Pandey and Abhishek Kadian and Ahmad Al-Dahle and Aiesha Letman and Akhil Mathur and Alan Schelten and Alex Vaughan and Amy Yang and Angela Fan and Anirudh Goyal and Anthony Hartshorn and Aobo Yang and Archi Mitra and Archie Sravankumar and Artem Korenev and Arthur Hinsvark and Arun Rao and Aston Zhang and Aurelien Rodriguez and Austen Gregerson and Ava Spataru and Baptiste Roziere and Bethany Biron and Binh Tang and Bobbie Chern and Charlotte Caucheteux and Chaya Nayak and Chloe Bi and Chris Marra and Chris McConnell and Christian Keller and Christophe Touret and Chunyang Wu and Corinne Wong and Cristian Canton Ferrer and Cyrus Nikolaidis and Damien Allonsius and Daniel Song and Danielle Pintz and Danny Livshits and Danny Wyatt and David Esiobu and Dhruv Choudhary and Dhruv Mahajan and Diego Garcia-Olano and Diego Perino and Dieuwke Hupkes and Egor Lakomkin and Ehab AlBadawy and Elina Lobanova and Emily Dinan and Eric Michael Smith and Filip Radenovic and Francisco Guzmán and Frank Zhang and Gabriel Synnaeve and Gabrielle Lee and Georgia Lewis Anderson and Govind Thattai and Graeme Nail and Gregoire Mialon and Guan Pang and Guillem Cucurell and Hailey Nguyen and Hannah Korevaar and Hu Xu and Hugo Touvron and Iliyan Zarov and Imanol Arrieta Ibarra and Isabel Kloumann and Ishan Misra and Ivan Evtimov and Jack Zhang and Jade Copet and Jaewon Lee and Jan Geffert and Jana Vranes and Jason Park and Jay Mahadeokar and Jeet Shah and Jelmer van der Linde and Jennifer Billock and Jenny Hong and Jenya Lee and Jeremy Fu and Jianfeng Chi and Jianyu Huang and Jiawen Liu and Jie Wang and Jiecao Yu and Joanna Bitton and Joe Spisak and Jongsoo Park and Joseph Rocca and Joshua Johnstun and Joshua Saxe and Junteng Jia and Kalyan Vasuden Alwala and Karthik Prasad and Kartikeya Upasani and Kate Plawiak and Ke Li and Kenneth Heafield and Kevin Stone and Khalid El-Arini and Krithika Iyer and Kshitiz Malik and Kuenley Chiu and Kunal Bhalla and Kushal Lakhotia and Lauren Rantala-Yeary and Laurens van der Maaten and Lawrence Chen and Liang Tan and Liz Jenkins and Louis Martin and Lovish Madaan and Lubo Malo and Lukas Blecher and Lukas Landzaat and Luke de Oliveira and Madeline Muzzi and Mahesh Pasupuleti and Mannat Singh and Manohar Paluri and Marcin Kardas and Maria Tsimpoukelli and Mathew Oldham and Mathieu Rita and Maya Pavlova and Melanie Kambadur and Mike Lewis and Min Si and Mitesh Kumar Singh and Mona Hassan and Naman Goyal and Narjes Torabi and Nikolay Bashlykov and Nikolay Bogoychev and Niladri Chatterji and Ning Zhang and Olivier Duchenne and Onur Çelebi and Patrick Alrassy and Pengchuan Zhang and Pengwei Li and Petar Vasic and Peter Weng and Prajjwal Bhargava and Pratik Dubal and Praveen Krishnan and Punit Singh Koura and Puxin Xu and Qing He and Qingxiao Dong and Ragavan Srinivasan and Raj Ganapathy and Ramon Calderer and Ricardo Silveira Cabral and Robert Stojnic and Roberta Raileanu and Rohan Maheswari and Rohit Girdhar and Rohit Patel and Romain Sauvestre and Ronnie Polidoro and Roshan Sumbaly and Ross Taylor and Ruan Silva and Rui Hou and Rui Wang and Saghar Hosseini and Sahana Chennabasappa and Sanjay Singh and Sean Bell and Seohyun Sonia Kim and Sergey Edunov and Shaoliang Nie and Sharan Narang and Sharath Raparthy and Sheng Shen and Shengye Wan and Shruti Bhosale and Shun Zhang and Simon Vandenhende and Soumya Batra and Spencer Whitman and Sten Sootla and Stephane Collot and Suchin Gururangan and Sydney Borodinsky and Tamar Herman and Tara Fowler and Tarek Sheasha and Thomas Georgiou and Thomas Scialom and Tobias Speckbacher and Todor Mihaylov and Tong Xiao and Ujjwal Karn and Vedanuj Goswami and Vibhor Gupta and Vignesh Ramanathan and Viktor Kerkez and Vincent Gonguet and Virginie Do and Vish Vogeti and Vítor Albiero and Vladan Petrovic and Weiwei Chu and Wenhan Xiong and Wenyin Fu and Whitney Meers and Xavier Martinet and Xiaodong Wang and Xiaofang Wang and Xiaoqing Ellen Tan and Xide Xia and Xinfeng Xie and Xuchao Jia and Xuewei Wang and Yaelle Goldschlag and Yashesh Gaur and Yasmine Babaei and Yi Wen and Yiwen Song and Yuchen Zhang and Yue Li and Yuning Mao and Zacharie Delpierre Coudert and Zheng Yan and Zhengxing Chen and Zoe Papakipos and Aaditya Singh and Aayushi Srivastava and Abha Jain and Adam Kelsey and Adam Shajnfeld and Adithya Gangidi and Adolfo Victoria and Ahuva Goldstand and Ajay Menon and Ajay Sharma and Alex Boesenberg and Alexei Baevski and Allie Feinstein and Amanda Kallet and Amit Sangani and Amos Teo and Anam Yunus and Andrei Lupu and Andres Alvarado and Andrew Caples and Andrew Gu and Andrew Ho and Andrew Poulton and Andrew Ryan and Ankit Ramchandani and Annie Dong and Annie Franco and Anuj Goyal and Aparajita Saraf and Arkabandhu Chowdhury and Ashley Gabriel and Ashwin Bharambe and Assaf Eisenman and Azadeh Yazdan and Beau James and Ben Maurer and Benjamin Leonhardi and Bernie Huang and Beth Loyd and Beto De Paola and Bhargavi Paranjape and Bing Liu and Bo Wu and Boyu Ni and Braden Hancock and Bram Wasti and Brandon Spence and Brani Stojkovic and Brian Gamido and Britt Montalvo and Carl Parker and Carly Burton and Catalina Mejia and Ce Liu and Changhan Wang and Changkyu Kim and Chao Zhou and Chester Hu and Ching-Hsiang Chu and Chris Cai and Chris Tindal and Christoph Feichtenhofer and Cynthia Gao and Damon Civin and Dana Beaty and Daniel Kreymer and Daniel Li and David Adkins and David Xu and Davide Testuggine and Delia David and Devi Parikh and Diana Liskovich and Didem Foss and Dingkang Wang and Duc Le and Dustin Holland and Edward Dowling and Eissa Jamil and Elaine Montgomery and Eleonora Presani and Emily Hahn and Emily Wood and Eric-Tuan Le and Erik Brinkman and Esteban Arcaute and Evan Dunbar and Evan Smothers and Fei Sun and Felix Kreuk and Feng Tian and Filippos Kokkinos and Firat Ozgenel and Francesco Caggioni and Frank Kanayet and Frank Seide and Gabriela Medina Florez and Gabriella Schwarz and Gada Badeer and Georgia Swee and Gil Halpern and Grant Herman and Grigory Sizov and Guangyi and Zhang and Guna Lakshminarayanan and Hakan Inan and Hamid Shojanazeri and Han Zou and Hannah Wang and Hanwen Zha and Haroun Habeeb and Harrison Rudolph and Helen Suk and Henry Aspegren and Hunter Goldman and Hongyuan Zhan and Ibrahim Damlaj and Igor Molybog and Igor Tufanov and Ilias Leontiadis and Irina-Elena Veliche and Itai Gat and Jake Weissman and James Geboski and James Kohli and Janice Lam and Japhet Asher and Jean-Baptiste Gaya and Jeff Marcus and Jeff Tang and Jennifer Chan and Jenny Zhen and Jeremy Reizenstein and Jeremy Teboul and Jessica Zhong and Jian Jin and Jingyi Yang and Joe Cummings and Jon Carvill and Jon Shepard and Jonathan McPhie and Jonathan Torres and Josh Ginsburg and Junjie Wang and Kai Wu and Kam Hou U and Karan Saxena and Kartikay Khandelwal and Katayoun Zand and Kathy Matosich and Kaushik Veeraraghavan and Kelly Michelena and Keqian Li and Kiran Jagadeesh and Kun Huang and Kunal Chawla and Kyle Huang and Lailin Chen and Lakshya Garg and Lavender A and Leandro Silva and Lee Bell and Lei Zhang and Liangpeng Guo and Licheng Yu and Liron Moshkovich and Luca Wehrstedt and Madian Khabsa and Manav Avalani and Manish Bhatt and Martynas Mankus and Matan Hasson and Matthew Lennie and Matthias Reso and Maxim Groshev and Maxim Naumov and Maya Lathi and Meghan Keneally and Miao Liu and Michael L. Seltzer and Michal Valko and Michelle Restrepo and Mihir Patel and Mik Vyatskov and Mikayel Samvelyan and Mike Clark and Mike Macey and Mike Wang and Miquel Jubert Hermoso and Mo Metanat and Mohammad Rastegari and Munish Bansal and Nandhini Santhanam and Natascha Parks and Natasha White and Navyata Bawa and Nayan Singhal and Nick Egebo and Nicolas Usunier and Nikhil Mehta and Nikolay Pavlovich Laptev and Ning Dong and Norman Cheng and Oleg Chernoguz and Olivia Hart and Omkar Salpekar and Ozlem Kalinli and Parkin Kent and Parth Parekh and Paul Saab and Pavan Balaji and Pedro Rittner and Philip Bontrager and Pierre Roux and Piotr Dollar and Polina Zvyagina and Prashant Ratanchandani and Pritish Yuvraj and Qian Liang and Rachad Alao and Rachel Rodriguez and Rafi Ayub and Raghotham Murthy and Raghu Nayani and Rahul Mitra and Rangaprabhu Parthasarathy and Raymond Li and Rebekkah Hogan and Robin Battey and Rocky Wang and Russ Howes and Ruty Rinott and Sachin Mehta and Sachin Siby and Sai Jayesh Bondu and Samyak Datta and Sara Chugh and Sara Hunt and Sargun Dhillon and Sasha Sidorov and Satadru Pan and Saurabh Mahajan and Saurabh Verma and Seiji Yamamoto and Sharadh Ramaswamy and Shaun Lindsay and Shaun Lindsay and Sheng Feng and Shenghao Lin and Shengxin Cindy Zha and Shishir Patil and Shiva Shankar and Shuqiang Zhang and Shuqiang Zhang and Sinong Wang and Sneha Agarwal and Soji Sajuyigbe and Soumith Chintala and Stephanie Max and Stephen Chen and Steve Kehoe and Steve Satterfield and Sudarshan Govindaprasad and Sumit Gupta and Summer Deng and Sungmin Cho and Sunny Virk and Suraj Subramanian and Sy Choudhury and Sydney Goldman and Tal Remez and Tamar Glaser and Tamara Best and Thilo Koehler and Thomas Robinson and Tianhe Li and Tianjun Zhang and Tim Matthews and Timothy Chou and Tzook Shaked and Varun Vontimitta and Victoria Ajayi and Victoria Montanez and Vijai Mohan and Vinay Satish Kumar and Vishal Mangla and Vlad Ionescu and Vlad Poenaru and Vlad Tiberiu Mihailescu and Vladimir Ivanov and Wei Li and Wenchen Wang and Wenwen Jiang and Wes Bouaziz and Will Constable and Xiaocheng Tang and Xiaojian Wu and Xiaolan Wang and Xilun Wu and Xinbo Gao and Yaniv Kleinman and Yanjun Chen and Ye Hu and Ye Jia and Ye Qi and Yenda Li and Yilin Zhang and Ying Zhang and Yossi Adi and Youngjin Nam and Yu and Wang and Yu Zhao and Yuchen Hao and Yundi Qian and Yunlu Li and Yuzi He and Zach Rait and Zachary DeVito and Zef Rosnbrick and Zhaoduo Wen and Zhenyu Yang and Zhiwei Zhao and Zhiyu Ma},
  journal={arXiv preprint arXiv:2407.21783},
  year={2024},
  url={https://arxiv.org/abs/2407.21783}
}

@inproceedings{arjona2019rudder,
 author = {Arjona-Medina, Jose A. and Gillhofer, Michael and Widrich, Michael and Unterthiner, Thomas and Brandstetter, Johannes and Hochreiter, Sepp},
 booktitle = {Advances in Neural Information Processing Systems},
 pages = {},
 publisher = {Curran Associates, Inc.},
 title = {RUDDER: Return Decomposition for Delayed Rewards},
 url = {https://proceedings.neurips.cc/paper_files/paper/2019/file/16105fb9cc614fc29e1bda00dab60d41-Paper.pdf},
 volume = {32},
 year = {2019}
}

@book{sutton2018reinforcement,
  title={Reinforcement Learning: An Introduction},
  author={Sutton, Richard S and Barto, Andrew G},
  year={2018},
  edition={Second},
  publisher={MIT Press}
}

@InProceedings{mnih2016asynchronous,
  title = 	 {Asynchronous Methods for Deep Reinforcement Learning},
  author = 	 {Mnih, Volodymyr and Badia, Adria Puigdomenech and Mirza, Mehdi and Graves, Alex and Lillicrap, Timothy and Harley, Tim and Silver, David and Kavukcuoglu, Koray},
  booktitle = 	 {Proceedings of The 33rd International Conference on Machine Learning},
  pages = 	 {1928--1937},
  year = 	 {2016},
  volume = 	 {48},
  series = 	 {Proceedings of Machine Learning Research},
  month = 	 {20--22 Jun},
  publisher =    {PMLR},
  url = 	 {https://proceedings.mlr.press/v48/mniha16.html}
}

@InProceedings{haarnoja2018soft,
  title = 	 {Soft Actor-Critic: Off-Policy Maximum Entropy Deep Reinforcement Learning with a Stochastic Actor},
  author =       {Haarnoja, Tuomas and Zhou, Aurick and Abbeel, Pieter and Levine, Sergey},
  booktitle = 	 {Proceedings of the 35th International Conference on Machine Learning},
  pages = 	 {1861--1870},
  year = 	 {2018},
  volume = 	 {80},
  series = 	 {Proceedings of Machine Learning Research},
  month = 	 {10--15 Jul},
  publisher =    {PMLR},
  url = 	 {https://proceedings.mlr.press/v80/haarnoja18b.html}
}

@inproceedings{ahmadian2024back,
    title = {Back to Basics: Revisiting {REINFORCE}-Style Optimization for Learning from Human Feedback in {LLM}s},
    author = {Ahmadian, Arash  and
      Cremer, Chris  and
      Gall{\'e}, Matthias  and
      Fadaee, Marzieh  and
      Kreutzer, Julia  and
      Pietquin, Olivier  and
      {\"U}st{\"u}n, Ahmet  and
      Hooker, Sara},
    booktitle = {Proceedings of the 62nd Annual Meeting of the Association for Computational Linguistics (Volume 1: Long Papers)},
    month = aug,
    year = {2024},
    address = {Bangkok, Thailand},
    publisher = {Association for Computational Linguistics},
    url = {https://aclanthology.org/2024.acl-long.662/},
    doi = {10.18653/v1/2024.acl-long.662},
    pages = {12248--12267}
}

@inproceedings{feng2025gigpo,
  title={Group-in-Group Policy Optimization for {LLM} Agent Training},
  author={Feng, Lang and Xue, Zhenghai and Liu, Tingcong and An, Bo},
  booktitle={Advances in Neural Information Processing Systems},
  pages={46375--46408},
  publisher={Curran Associates, Inc.},
  url={https://proceedings.neurips.cc/paper_files/paper/2025/file/420c9f777c0b4f78d515e53cf74d58b2-Paper-Conference.pdf},
  volume={38},
  year={2025}
}

@inproceedings{li2025salt,
    title = {{SALT}: Step-level Advantage Assignment for Long-horizon Agents via Trajectory Graph},
    author = {Li, Jiazheng  and
      Wang, Yawei  and
      Yan, Qiaojing  and
      Tian, Yijun  and
      Xu, Zhichao  and
      Song, Huan  and
      Xu, Panpan  and
      Cheong, Lin Lee},
    booktitle = {Findings of the {A}ssociation for {C}omputational {L}inguistics: {EACL} 2026},
    month = mar,
    year = {2026},
    address = {Rabat, Morocco},
    publisher = {Association for Computational Linguistics},
    url = {https://aclanthology.org/2026.findings-eacl.247/},
    doi = {10.18653/v1/2026.findings-eacl.247},
    pages = {4709--4725},
}

@article{jin2025search,
  title={Search-r1: Training llms to reason and leverage search engines with reinforcement learning},
  author={Jin, Bowen and Zeng, Hansi and Yue, Zhenrui and Yoon, Jinsung and Arik, Sercan and Wang, Dong and Zamani, Hamed and Han, Jiawei},
  journal={arXiv preprint arXiv:2503.09516},
  year={2025}
}

@inproceedings{perez2020unsupervised,
    title = {Unsupervised Question Decomposition for Question Answering},
    author = {Perez, Ethan  and
      Lewis, Patrick  and
      Yih, Wen-tau  and
      Cho, Kyunghyun  and
      Kiela, Douwe},
    booktitle = {Proceedings of the 2020 Conference on Empirical Methods in Natural Language Processing (EMNLP)},
    month = nov,
    year = {2020},
    address = {Online},
    publisher = {Association for Computational Linguistics},
    url = {https://aclanthology.org/2020.emnlp-main.713/},
    doi = {10.18653/v1/2020.emnlp-main.713},
    pages = {8864--8880},
}

@inproceedings{khot2022decompose,
  title={Decomposed Prompting: A Modular Approach for Solving Complex Tasks},
  author={Tushar Khot and Harsh Trivedi and Matthew Finlayson and Yao Fu and Kyle Richardson and Peter Clark and Ashish Sabharwal},
  booktitle={The Eleventh International Conference on Learning Representations},
  year={2023},
  url={https://openreview.net/forum?id=_nGgzQjzaRy}
}

@inproceedings{khattab2021baleen,
 author = {Khattab, Omar and Potts, Christopher and Zaharia, Matei},
 booktitle = {Advances in Neural Information Processing Systems},
 pages = {27670--27682},
 publisher = {Curran Associates, Inc.},
 title = {Baleen: Robust Multi-Hop Reasoning at Scale via Condensed Retrieval},
 url = {https://proceedings.neurips.cc/paper_files/paper/2021/file/e8b1cbd05f6e6a358a81dee52493dd06-Paper.pdf},
 volume = {34},
 year = {2021}
}

@inproceedings{qi2021answering,
    title = {Answering Open-Domain Questions of Varying Reasoning Steps from Text},
    author = {Qi, Peng  and
      Lee, Haejun  and
      Sido, Tg  and
      Manning, Christopher},
    booktitle = {Proceedings of the 2021 Conference on Empirical Methods in Natural Language Processing},
    month = nov,
    year = {2021},
    address = {Online and Punta Cana, Dominican Republic},
    publisher = {Association for Computational Linguistics},
    url = {https://aclanthology.org/2021.emnlp-main.292/},
    doi = {10.18653/v1/2021.emnlp-main.292},
    pages = {3599--3614}
}

@inproceedings{asai2022evidentiality,
    title = {Evidentiality-guided Generation for Knowledge-Intensive {NLP} Tasks},
    author = {Asai, Akari  and
      Gardner, Matt  and
      Hajishirzi, Hannaneh},
    booktitle = {Proceedings of the 2022 Conference of the North American Chapter of the Association for Computational Linguistics: Human Language Technologies},
    month = jul,
    year = {2022},
    address = {Seattle, United States},
    publisher = {Association for Computational Linguistics},
    url = {https://aclanthology.org/2022.naacl-main.162/},
    doi = {10.18653/v1/2022.naacl-main.162},
    pages = {2226--2243}
}

\appendix
 
\clearpage  

\section{Prompt Template}
\label{sec:appendix_prompt}

To ensure the backbone Large Language Models properly format their multi-step reasoning and tool-use procedures, we use a structured system prompt.
As illustrated in Figure~\ref{fig:prompt_template}, the prompt enforces a strict thought-action observation loop.
The agent is required to first output its reasoning traces within the \texttt{<think>} tags, followed by either a retrieval action (\texttt{<tool\_call>}) or a final answer generation (\texttt{<answer>}).
Both outputs are strictly constrained to a JSON list format to facilitate robust parsing.

\begin{figure}[h] 
\centering
\begin{tcolorbox}[
    colback=gray!5!white,
    colframe=gray!75!black,
    title=System Prompt Template, 
    arc=2mm,
    boxrule=0.5pt,
    left=1mm, right=1mm 
]
\fontsize{7.5pt}{8.5pt}\selectfont 
\begin{verbatim}
Answer the question. You must strictly follow this 
loop for every step:
1. Always start with reasoning inside <think>...</think>.
2. Then, output either a search tool call OR the 
   final answer.
   - To search: <tool_call>["query1", ...]</tool_call>
     (System returns top-5 snippets per query)
   - To answer: <answer>["A", "B"]</answer>

Both <tool_call> and <answer> content must be 
a JSON list of strings.

Question: {Question here}
\end{verbatim}
\end{tcolorbox}
\caption{The standard system prompt template.}
\label{fig:prompt_template}
\end{figure}

\section{Datasets}
\label{sec:appendix_Datasets}

To rigorously evaluate the diverse exploration and complex reasoning capabilities of the SPADER framework, we select four widely recognized benchmarks. Each dataset presents unique challenges in terms of logical constraints, long-tail entity retrieval, and multi-hop dependency:

\begin{compactitem}
    \item \textbf{QAMPARI} \cite{amouyal2023qampari}: An open-domain question answering (ODQA) benchmark specifically designed for questions that have many answers spread across multiple paragraphs. Unlike standard datasets that focus on a single answer from a single paragraph, every question in QAMPARI has at least 5 answers, averaging 13 answers per question. It tests a system's ability to retrieve and read a large number of passages from a corpus to generate exhaustive lists of entities.
    \item \textbf{Mintaka} \cite{sen2022mintaka}: A large-scale, multilingual question answering dataset containing 20,000 question-answer pairs elicited in English and translated into eight other languages. A key differentiator is its focus on naturalness; rather than auto-generating queries, it features complex questions naturally elicited from crowd workers. It spans 8 distinct complexity types: count, comparative, superlative, ordinal, multi-hop, intersection, difference, and yes/no questions.
    \item \textbf{WebQSP} \cite{yih2016value}: A Knowledge Base Question Answering (KBQA) dataset that provides full SPARQL semantic parses for 4,737 questions from the original WebQuestions dataset. Rather than being designed for open-book text retrieval, it is explicitly grounded in Freebase to evaluate the value of training models on intermediate logical forms (semantic parses) rather than just final question-answer pairs.
    \item \textbf{QUEST} \cite{malaviya2023quest}: A retrieval benchmark comprising 3,357 natural language queries that test a model's ability to handle implicit set operations, such as intersection, union, and difference. Each query maps to an exhaustive answer set of 2 to 20 Wikipedia entities. It challenges retrieval systems to match multiple constraints mentioned in a query with corresponding textual evidence in documents, without relying on structured knowledge bases.
\end{compactitem}

\section{Implementation Details}
\label{sec:appendix_implementation}

\paragraph{Framework and Infrastructure.}
Our training pipeline is built upon the Verl framework~\cite{sheng2024hybridflow} for Reinforcement Learning optimization, utilizing sglang as the high-throughput inference engine for both training-time rollout sampling and evaluation.
All experiments are conducted on a single node equipped with 8 NVIDIA A100 GPUs. 

\paragraph{Training Configurations and Hyperparameters.}
The unified training set consists of 12,000 instances, including 6,000 from QAMPARI and 2,000 each from Mintaka, WebQSP, and QUEST.
We train the models for 1 epoch, resulting in 375 optimization steps with a global batch size of 32.
For gradient stability, we set the mini-batch size to 32 and micro-batch size to 4.
We employ the AdamW optimizer with a learning rate of 1e-5 and a linear warmup ratio of 0.05.
To facilitate step-wise policy optimization, our SPADER framework samples a group of $G=16$ parallel trajectories for each prompt.
To maintain policy stability, the KL-divergence penalty coefficient $\beta_{\mathrm{KL}}$ is fixed at 0.001.
For advantage estimation, We set the discount factor $\gamma$ to 0.9 to balance immediate rewards and long-term credit assignment. 

\paragraph{Reward and Generation Settings.}
During the reinforcement learning phase, we use a temperature of 1.0 and \textit{top\_p} of 1.0 for diverse trajectory generation.
The Diversity-Aware Exploration Reward is configured with a horizontal information gain weight $\alpha = 1.0$ and a vertical novelty premium weight $\beta = 4.0$.
To prevent redundant tool calls and encourage efficiency, a tool-use cost of 0.01 is applied per search call.

\paragraph{Entity Normalization and Step-level Matching.}
For step-level reward settlement, we process the complete trajectory after each rollout.
At decision step $t$, we take the returned tool-response snippets and determine step-level hits by matching the ground-truth entity set against these snippets after deterministic normalization, including lowercasing, Unicode normalization, punctuation and article removal, and whitespace collapse.
Newly discovered entities at step $t$ are the step-level hits that are not in the cumulative matched set from earlier steps.

We acknowledge that this string-based protocol may miss some lexical variants.
However, this mainly introduces false negatives, so the induced bias primarily lowers absolute step-level rewards rather than inflating them.
At the same time, because the same matching procedure is applied to all trajectories within a group, these misses are often shared across peers and therefore have limited effect on step-wise relative comparison under the group baseline.
More robust matching remains an important direction for future investigation.

\section{Full Ablation Results}
\label{sec:appendix_ablation}

For completeness, we report the full ablation results with Precision (P), Recall (R), and F1 across all four datasets.

\begin{table*}[!t]
\small
\centering
\renewcommand{\arraystretch}{1.4}
\tabcolsep=2pt
\begin{tabularx}{\textwidth}{@{} l *{12}{Y} @{}}
\toprule
& \multicolumn{3}{c}{\textbf{QAMPARI}} & \multicolumn{3}{c}{\textbf{Mintaka}} & \multicolumn{3}{c}{\textbf{WebQSP}} & \multicolumn{3}{c}{\textbf{QUEST}} \\ \cmidrule(l){2-13}
\multirow{-2}{*}{\textbf{Methods}} & P & R & F1 & P & R & F1 & P & R & F1 & P & R & F1 \\ \midrule
SPADER & \textbf{0.474} & \textbf{0.312} & \textbf{0.343} & \textbf{0.625} & \textbf{0.626} & \textbf{0.611} & \textbf{0.423} & \textbf{0.409} & \textbf{0.388} & \textbf{0.233} & 0.109 & \textbf{0.127} \\
w/o SPA & 0.404 & 0.268 & 0.285 & 0.591 & 0.566 & 0.569 & 0.380 & 0.343 & 0.336 & 0.184 & \textbf{0.110} & 0.117 \\
w/o Novelty Premium & 0.309 & 0.259 & 0.239 & 0.344 & 0.505 & 0.373 & 0.250 & 0.341 & 0.256 & 0.092 & 0.095 & 0.083 \\
w/o Info Gain & 0.399 & 0.224 & 0.266 & 0.340 & 0.425 & 0.364 & 0.345 & 0.414 & 0.359 & 0.164 & 0.074 & 0.094 \\ \bottomrule
\end{tabularx}
\caption{Full ablation results of SPADER on Qwen3-8B. ``w/o'' denotes removing one component from SPADER.}
\label{tab:ablation-full}
\end{table*}

\section{Case Study}
\label{app:case_study}

This section analyzes QAMPARI trajectories sampled from the Qwen3-8B backbone.
Cases 1--3 use the same prompt to compare ReAct, SPADER, and the SPADER ablation without the information gain.
Case 4 uses a different prompt to illustrate an ambiguity-driven mismatch between natural-language intent and evaluation target.

\paragraph{Case 1 and Case 2: Early Stop vs. Iterative Expansion.}
Case 1 and Case 2 share the same 75-answer prompt on Luzerne County municipalities, as shown in Figure~\ref{fig:case_baseline_early_stop} and Figure~\ref{fig:case_ours_incremental}.
In Case 1, ReAct exposes a useful decomposition signal in the first search action, including city, borough, and township cues, but terminates immediately after three calls and returns only 7 entities.
Precision is not extremely poor, yet recall drops to 0.03, indicating a typical under-exploration pattern where coarse evidence is not converted into broader coverage.

In Case 2, SPADER does not only use more search actions, but also organizes them with a clearer progression.
Step 1 builds a seed set, Step 2 branches by municipality type, and Step 3 shifts to a completeness-oriented sweep.
Although later snippets become noisier, the accumulated entity set still expands from 7 to 42 predicted items, and F1 rises from 0.05 to 0.24.

\paragraph{Case 3: Ablation Dynamics Without Information Gain.}
Case 3 is shown in Figure~\ref{fig:case_ablation_no_infogain}.
After removing the information gain component from the Diversity-Aware Exploration Reward, the early phase still discovers valid municipality names.
However, after this short productive window, the trajectory enters a long low-yield regime in which many additional search actions produce almost no new entities.

The trace suggests a specific failure pattern: query templates keep changing lexically, while returned evidence remains semantically repetitive, dominated by county-level background text and previously seen names.
Without a strong incentive for incremental useful gain, the policy becomes less sensitive to diminishing returns and delays termination.
The final outcome, 34 calls with 5 predictions and F1 of 0.05, is more consistent with search-action drift plus poor stopping than with a single catastrophic tool error.

\paragraph{Case 4: Ambiguity Under Broad Intent.}
Case 4 is shown in Figure~\ref{fig:case_ours_failure_insight}. The prompt ``Who served the country of Australia?'' is semantically broad and can naturally be interpreted as political leaders or senior public officials, while the benchmark target set follows a different entity scope.
Under this ambiguity, the first search action anchors the trajectory to a leadership interpretation that is semantically plausible but label-misaligned.

The resulting zero-overlap outcome should be interpreted cautiously.
It reflects both policy behavior and a task-specification mismatch between natural-language intent and evaluation target.
In this example, additional search depth, 9 calls and 37 predicted entities, does not resolve the mismatch once the initial interpretation diverges from the annotation scope.

\paragraph{Overall Takeaway.}
Across these examples, we observe two practical patterns.
When intermediate evidence can be transformed into targeted follow-up search actions, SPADER tends to improve coverage over shallow one-step behavior.
At the same time, some low-score cases also show a clear boundary, when prompt intent is broad or ambiguous, early intent misalignment can dominate the final error.

\clearpage
\begin{figure*}[t]
\centering
\begin{tcolorbox}[
  title=ReAct: Early Stop,
  colback=red!2!white,
  colframe=red!50!black,
  boxrule=0.6pt,
  sharp corners,
  fonttitle=\bfseries,
  width=0.98\linewidth,
  enhanced, segmentation style={draw=none}
]
\textbf{Prompt:} \emph{What municipalities are located in Luzerne County, Pennsylvania?}

\textbf{Ground Truth:} total=75. Sample: ["Wilkes-Barre city", "Hazleton city", "Kingston borough", "Hanover Township", "Nanticoke city", "Plains Township", "Hazle Township", "Butler Township", ...].
\begin{tcolorbox}[colback=white, colframe=black!30, boxrule=0.4pt, sharp corners, breakable, enhanced, segmentation style={draw=none}]
\footnotesize
\textbf{Step 1}\\
\textbf{search:} ["municipalities in Luzerne County Pennsylvania", "official list of municipalities Luzerne County PA", "Luzerne County PA cities towns boroughs"]\\
\textbf{tool response:} [
"Luzerne County ... contains one of the highest numbers of independently governed municipalities in Pennsylvania ...";
"Bear Creek Township is a township in Luzerne County, Pennsylvania ...";
"Plains Township is a township in Luzerne County, Pennsylvania ..."; etc.
]\\
\textbf{new entities:} ["Bear Creek Township", "Plains Township", "Wilkes-Barre", "Pittston", ...]\\
\textbf{next-step rationale:} county-level list should be expanded by municipality type, but the trajectory ends early.
\end{tcolorbox}

\begin{tcolorbox}[colback=white, colframe=black!30, boxrule=0.4pt, sharp corners, breakable, enhanced, segmentation style={draw=none}]
\footnotesize
\textbf{Step 2 }\\
\textbf{answer:} \textcolor{red!70!black}{\textbf{[Final Answer]}} ["Wilkes-Barre", "Pittston", "Harveys Lake", "Jeddo", "Bear Creek", "Wilkes-Barre Township", "Plains Township"]
\end{tcolorbox}

\textbf{Metadata:} predicted\_count=7, precision=0.29, recall=0.03, f1=0.05, call\_count=3.
\end{tcolorbox}
\caption{ReAct: early stop after one search round leads to severe under-coverage.}
\label{fig:case_baseline_early_stop}
\end{figure*}

\begin{figure*}[t]
\centering
\begin{tcolorbox}[
  title=SPADER: Iterative Expansion with Information Gain,
  colback=green!2!white,
  colframe=green!45!black,
  boxrule=0.6pt,
  sharp corners,
  fonttitle=\bfseries,
  width=0.98\linewidth,
  enhanced, segmentation style={draw=none}
]
\textbf{Prompt:} \emph{What municipalities are located in Luzerne County, Pennsylvania?}

\textbf{Ground Truth:} total=75. Sample: ["Wilkes-Barre city", "Hazleton city", "Kingston borough", "Hanover Township", "Nanticoke city", "Plains Township", "Hazle Township", "Butler Township", ...].
\begin{tcolorbox}[colback=white, colframe=black!30, boxrule=0.4pt, sharp corners, breakable, enhanced, segmentation style={draw=none}]
\footnotesize
\textbf{Step 1}\\
\textbf{search:} ["list of towns in Luzerne County PA", "municipalities Luzerne County Pennsylvania", "cities boroughs Luzerne PA"]\\
\textbf{tool response:} [
"Plains Township is a township in Luzerne County, Pennsylvania ...";
"Hazle Township is a township in Luzerne County, Pennsylvania ...";
"Luzerne County ... has a very large number of municipalities ..."; etc.
]\\
\textbf{new entities:} ["Hazle Township", "Plains Township", "Bear Creek Township", "Wilkes-Barre", "Pittston", "Hazleton", ...]\\
\textbf{next-step rationale:} first-hop evidence is township-heavy; next search should explicitly decompose by \emph{cities / boroughs / townships}.
\end{tcolorbox}

\begin{tcolorbox}[colback=white, colframe=black!30, boxrule=0.4pt, sharp corners, breakable, enhanced, segmentation style={draw=none}]
\footnotesize
\textbf{Step 2}\\
\textbf{search:} ["list of cities in Luzerne County PA", "boroughs in Luzerne County Pennsylvania 2023", "townships Luzerne County Pennsylvania names"]\\
\textbf{tool response:} [
"Plains Township ... township on the outskirts of Wilkes-Barre ...";
"Hazle Township ... township in Luzerne County ...";
"Luzerne County ... municipalities by local-government type ..."; etc.
]\\
\textbf{new entities:} ["Foster Township", "Butler Township", "Conyngham", "Sugarloaf Township", "Black Creek Township", "Harveys Lake", "Jeddo", ...]\\
\textbf{next-step rationale:} type-specific queries recover many new names; now switch to completeness-oriented township inventory and by-type consolidation.
\end{tcolorbox}

\begin{tcolorbox}[colback=white, colframe=black!30, boxrule=0.4pt, sharp corners, breakable, enhanced, segmentation style={draw=none}]
\footnotesize
\textbf{Step 3}\\
\textbf{search:} ["list of townships in Luzerne County PA 2023", "municipalities in Luzerne County PA by type", "townships in Luzerne County Pennsylvania with names"]\\
\textbf{tool response:} [
"Luzerne County ... administrative history and county-level references ...";
"state of Pennsylvania and the county seat of Luzerne County ...";
"Luzerne County ... large municipal count indicates remaining uncovered entities ..."; etc.
]\\
\textbf{new entities:} ["Dallas Township", "Huntington Township", "Franklin Township", "Jackson Township", "Lake Township", "Lehman Township", ...]\\
\textbf{next-step rationale:} despite noisier snippets, prior accumulated entity set keeps growing and supports a broader final consolidation.
\end{tcolorbox}

\begin{tcolorbox}[colback=white, colframe=black!30, boxrule=0.4pt, sharp corners, breakable, enhanced, segmentation style={draw=none}]
\footnotesize
\textbf{Step 4 }\\
\textbf{answer:} \textcolor{green!45!black}{\textbf{[Final Answer]}} ["Hazle Township", "Plains Township", "Bear Creek Township", "Wilkes-Barre", "Pittston", "Hazleton", "Luzerne", "Kingston", "Nanticoke", "West Hazleton", "Foster Township", "Butler Township", "Conyngham", "Sugarloaf Township", "Black Creek Township", "Harveys Lake", "Jeddo", "Wilkes-Barre Township", "Dallas Township", "Huntington Township", "Franklin Township", "Jackson Township", "Lake Township", "Lehman Township", ...]
\end{tcolorbox}

\textbf{Metadata:} predicted\_count=42, precision=0.33, recall=0.19, f1=0.24, call\_count=9.
\end{tcolorbox}
\caption{SPADER: incremental exploration expands coverage step by step.}
\label{fig:case_ours_incremental}
\end{figure*}

\begin{figure*}[t]
\centering
\begin{tcolorbox}[
  title=SPADER without Information Gain: From Useful Start to Search-Action Drift,
  colback=yellow!3!white,
  colframe=orange!65!black,
  boxrule=0.6pt,
  sharp corners,
  fonttitle=\bfseries,
  width=0.98\linewidth,
  enhanced, segmentation style={draw=none}
]
\textbf{Prompt:} \emph{What municipalities are located in Luzerne County, Pennsylvania?}

\textbf{Ground Truth:} total=75. Sample: ["Wilkes-Barre city", "Hazleton city", "Kingston borough", "Hanover Township", "Nanticoke city", "Plains Township", "Hazle Township", "Butler Township", ...].
\begin{tcolorbox}[colback=white, colframe=black!30, boxrule=0.4pt, sharp corners, breakable, enhanced, segmentation style={draw=none}]
\footnotesize
\textbf{Step 1}\\
\textbf{search:} ["What municipalities are in Luzerne County PA", "List of towns in Luzerne County Pennsylvania"]\\
\textbf{tool response:} [
"Luzerne County ... has an exceptionally high number of municipalities ...";
"Bear Creek Township is a township in Luzerne County ...";
"Plains Township is a township in Luzerne County ..."; etc.
]\\
\textbf{new entities:} ["Bear Creek Township", "Plains Township", "Dallas", ...]\\
\textbf{next-step rationale:} early responses contain valid municipality clues.
\end{tcolorbox}

\begin{tcolorbox}[colback=white, colframe=black!30, boxrule=0.4pt, sharp corners, breakable, enhanced, segmentation style={draw=none}]
\footnotesize
\textbf{Step 2--4 (transition)}\\
\textbf{search:} ["List of cities and boroughs in Luzerne County PA", "Municipalities in Luzerne County Pennsylvania 2023"]; ["Luzerne County cities list PA", "Luzerne County boroughs PA"]; ["Luzerne County townships PA", "Luzerne County towns PA"]\\
\textbf{tool response:} [
"Dallas is a borough in Luzerne County, Pennsylvania ...";
"Bear Creek Township is a township in Luzerne County ...";
"County-level background lines increasingly repeat with limited new entities ..."; etc.
]\\
\textbf{new entities:} ["Dallas", "Newport Township", "Huntington Township", ...]\\
\textbf{next-step rationale:} signal begins to dilute; many search calls return county facts rather than new municipality names.
\end{tcolorbox}

\begin{tcolorbox}[colback=white, colframe=black!30, boxrule=0.4pt, sharp corners, breakable, enhanced, segmentation style={draw=none}]
\footnotesize
\textbf{Step 5--16 (drift / noisy exploration phase)}\\
\textbf{search:} ["Luzerne County municipalities PA 2023", "List of all towns in Luzerne County PA"]; ["Luzerne County municipal list 2023", "PA Luzerne County incorporated areas"]; ["Luzerne County city and borough list", "List of all Luzerne County townships"]; ["Luzerne County municipal list final 2023", "PA Luzerne County all incorporated areas complete"]; ...\\
\textbf{tool response:} [
"Huntington Township is a township in Luzerne County ...";
"Dallas Township is a township in Luzerne County ...";
"Repeated county-level boilerplate dominates (e.g., topography/history) with weak incremental gain ..."; etc.
]\\
\textbf{new entities:} [none]\\
\textbf{next-step rationale:} query templates keep mutating but mostly recycle existing evidence and fail to converge.
\end{tcolorbox}

\begin{tcolorbox}[colback=white, colframe=black!30, boxrule=0.4pt, sharp corners, breakable, enhanced, segmentation style={draw=none}]
\footnotesize
\textbf{Step 17 }\\
\textbf{answer:} \textcolor{orange!70!black}{\textbf{[Final Answer]}} ["Wilkes-Barre", "Hazleton", "Plains Township", "Bear Creek Township", "Dallas"]
\end{tcolorbox}

\textbf{Metadata:} predicted\_count=5, precision=0.40, recall=0.03, f1=0.05, call\_count=34.
\end{tcolorbox}
\caption{SPADER without information gain: search actions drift into repetitive, low-gain calls with degraded stopping behavior.}
\label{fig:case_ablation_no_infogain}
\end{figure*}

\begin{figure*}[t]
\centering
\begin{tcolorbox}[
  title=SPADER Failure Case: Broad Intent Causes Entity-Type Drift,
  colback=blue!2!white,
  colframe=blue!45!black,
  boxrule=0.6pt,
  sharp corners,
  fonttitle=\bfseries,
  width=0.98\linewidth,
  enhanced, segmentation style={draw=none}
]
\textbf{Prompt:} \emph{Who served the country of Australia?}

\textbf{Ground Truth:} total=63. Sample: ["John Murray", "Cecil Foott", "Dominic McCarthy", "Edmund Lind", "Jack Studdert", "Ringer Edwards", "Maurice Fergusson", "David Johnston", ...].
\begin{tcolorbox}[colback=white, colframe=black!30, boxrule=0.4pt, sharp corners, breakable, enhanced, segmentation style={draw=none}]
\footnotesize
\textbf{Step 1}\\
\textbf{search:} ["list of Australian Prime Ministers", "Governors-General of Australia list", "people who served as leaders in Australia government"]\\
\textbf{tool response:} [
"List of prime ministers of Australia ...";
"Historical rankings of prime ministers of Australia ...";
"Records of prime ministers of Australia ..."; etc.
]\\
\textbf{new entities:} ["Edmund Barton", "Andrew Fisher", "Joseph Cook", "John Curtin", ...]\\
\textbf{next-step rationale:} after a leader-heavy first pass, the model broadens query scope to additional public-service roles.
\end{tcolorbox}

\begin{tcolorbox}[colback=white, colframe=black!30, boxrule=0.4pt, sharp corners, breakable, enhanced, segmentation style={draw=none}]
\footnotesize
\textbf{Step 2}\\
\textbf{search:} ["list of Australian state premiers", "Australian Chief Justices list", "notable Australian public servants in government"]\\
\textbf{tool response:} [
"List of West Australian Football League premiers ...";
"West Australian State Premiership ...";
"Premier of Tasmania ..."; etc.
]\\
\textbf{new entities:} [none]\\
\textbf{next-step rationale:} retrieved content drifts away from target entity type, but the model continues broad government-style queries.
\end{tcolorbox}

\begin{tcolorbox}[colback=white, colframe=black!30, boxrule=0.4pt, sharp corners, breakable, enhanced, segmentation style={draw=none}]
\footnotesize
\textbf{Step 3}\\
\textbf{search:} ["list of Australian Cabinet Secretaries", "notable Australian public officials in government history", "senior public servants who served in Australian federal government"]\\
\textbf{tool response:} [
"List of West Australian Football League premiers ...";
"West Australian State Premiership ...";
"Premier of Tasmania ..."; etc.
]\\
\textbf{new entities:} [none]\\
\textbf{next-step rationale:} repeated off-target evidence is accumulated, and the model consolidates a leader-centric final list.
\end{tcolorbox}

\begin{tcolorbox}[colback=white, colframe=black!30, boxrule=0.4pt, sharp corners, breakable, enhanced, segmentation style={draw=none}]
\footnotesize
\textbf{Step 4 }\\
\textbf{answer:} \textcolor{blue!60!black}{\textbf{[Final Answer]}} ["Edmund Barton", "Andrew Fisher", "Joseph Cook", "John Curtin", "Ben Chifley", "Gough Whitlam", "Bob Hawke", "Paul Keating", "John Howard", "Kevin Rudd", ...]
\end{tcolorbox}

\textbf{Metadata:} predicted\_count=37, precision=0.00, recall=0.00, f1=0.00, call\_count=9.
\end{tcolorbox}
\caption{broad-intent ambiguity: a plausible but misaligned interpretation leads to low overlap with the target set.}
\label{fig:case_ours_failure_insight}
\end{figure*}

\end{document}